\documentclass{article}

    \usepackage[preprint]{neurips_2022}

\usepackage[utf8]{inputenc} 
\usepackage[T1]{fontenc}    
\usepackage{hyperref}       
\usepackage{url}            
\usepackage{booktabs}       
\usepackage{amsfonts}       
\usepackage{nicefrac}       
\usepackage{microtype}      
\usepackage{xcolor}         

\usepackage[pdftex]{graphicx}
\usepackage{caption}
\usepackage{subcaption}
\usepackage{hyperref}
\hypersetup{
    colorlinks=true,
    citecolor=black,
    linkcolor=black,
    urlcolor=blue,
    }
\usepackage{algorithm}
\usepackage{algpseudocode}
\usepackage{amsmath}
\DeclareMathOperator*{\argmax}{argmax} 
    
\usepackage{tikz}  
\usetikzlibrary{calc}
\tikzstyle{stuff_fill}=[circle,draw,fill=white,inner sep=0.4pt] 

\mathchardef\mhyphen="2D

\newcommand{\plmi}{\tiny$\pm$}

\title{muNet: Evolving Pretrained Deep Neural Networks \\ 
into Scalable Auto-tuning  
Multitask Systems}

\author{%
  Andrea Gesmundo \\
  Google Research \\
  \texttt{agesmundo@google.com} \\
  \And
  Jeff Dean \\
  Google Research \\
  \texttt{jeff@google.com} \\
}

\begin{document}

\maketitle

\begin{abstract}
Most uses of machine learning today 
involve training a model from scratch for a particular task,
or sometimes starting with a model pretrained on a related task and then fine-tuning on a downstream task.
Both approaches offer limited knowledge transfer between different tasks,
time-consuming human-driven customization to individual tasks and high computational costs especially when starting from randomly initialized models.
We propose a method that uses the layers of a pretrained deep neural network as building blocks to construct an ML system that can jointly solve an arbitrary number of tasks.
The resulting system can leverage cross tasks knowledge transfer, while being immune from common drawbacks of multitask approaches such as catastrophic forgetting, gradients interference and negative transfer.
We define an evolutionary approach designed %
to jointly select the prior knowledge relevant for each task, choose the subset of the model parameters to train and dynamically auto-tune its hyperparameters.
Furthermore, a novel scale control method is employed to achieve quality/size trade-offs that outperform common fine-tuning techniques.
Compared with standard fine-tuning on a benchmark of 10 diverse image classification tasks, the proposed model improves the average accuracy by 2.39\% while using 47\% less parameters per task. %

\end{abstract}

\section{Introduction}

ML techniques are increasingly successful in a growing number of applications, either by iteratively improving the state-of-the-art in impactful domains such as language \citep{Brown2020LanguageMA} 
and vision \citep{Dosovitskiy2021AnII}, or achieving new capabilities such as protein folding \citep{Senior2020ImprovedPS}, chip design \citep{Mirhoseini2020ChipPW}, 
superhuman performance in different competitions \citep{Silver2016MasteringTG, Vinyals2019GrandmasterLI}.
Although successful, the standard ML methodology is based on practices that 
limit 
the quality and efficiency of the produced solutions.
Some of these practices include:

\textbf{Single task models} \ \ The 
majority of ML practice, both in applications and research, aims to produce models that can solve a single task.
Such models can be customized and tuned to the task at hand, 
and in some cases 
achieve state-of-the-art results with a self-contained and well-defined methodology.

\textbf{Limited prior knowledge reuse} \ \ A significant portion of all ML models are trained from a random initialized state.
This inefficiency has been alleviated by the increased availability of reference pretrained models that can be used as a starting point for fine-tuning to produce a dedicated model for any target tasks with matching 
input modalities and task framing  \citep{Devlin2019BERTPO,Raffel2020ExploringTL}. 
However, the approach of training large base models, and then fine-tuning a separate copy of it 
on each downstream task,
loses out on the potential benefits of incorporating knowledge of the downstream tasks  
into the core model and enabling this knowledge to be reused for related tasks.

\textbf{Manual Tuning} \ \ The standard process of training an ML model
requires repeating the 
training multiple times to tune its hyperparameters and identify the configuration that yields the better results.

\textbf{Engineering efficiency} \ \ Traditional large-scale software systems enable teams of hundreds or thousands
of software engineers to work collectively on a single software artifact, through decomposition of the problem into many sub-problems, and through well-defined abstraction boundaries.
We currently lack the ability to have thousands of ML engineers and researchers collectively contribute to a single model.
By enabling automatic incorporation of new tasks and knowledge into a single running system, through evolutionary exploration, we see a direction where many people can all contribute to the improvement of a single overall ML model that is suited to a growing number of tasks, and that incorporates the learning and knowledge facilitated by many other people working on the same system.
In addition to enabling thousands of engineers and researchers to contribute to a single system, 
it may even be possible to have tens of millions of people without knowledge of ML training to contribute to training a single ML model, by contributing new tasks and examples and building on the skills that have been taught to the model by others.

We propose a method designed to explore
the identified opportunities for improvement of the standard ML methodology.
This method can jointly solve multiple tasks to achieve increased efficiency and quality for each.
The knowledge learned from each task is compartmentalized in components that can be reused by multiple tasks.
As the system accumulates the ability to solve 
more tasks, it is able to find better solutions for subsequent tasks, and to do so with increasing efficiency, requiring fewer added parameters for new tasks.
The knowledge compartmentalization allows to avoid common problems of multitask models such as catastrophic forgetting.
The exploration of model architectures and identification of the subset of prior knowledge most relevant for each task is guided by an 
evolutionary algorithm designed to dynamically adjust the exploration/exploitation balance without need of manual tuning of meta-parameters.
The same evolutionary logic is also employed to dynamically tune the hyperparameters of the components of the multitask model.
The proposed auto-tuning approach identifies a schedule of values over time for each hyperparameter rather than a single value.
We apply the proposed method to the domain of image classification, demonstrating empirically that it can achieve quality/size trade-offs that outperform common fine-tuning techniques. 
Furthermore, the proposed method can use of any pretrained model as a starting point of the evolution.
Allowing to 
build on top of prior work and further increasing efficiency in terms of convergence time.

\begin{figure}[t]
\centering
\includegraphics[width=1.\linewidth]{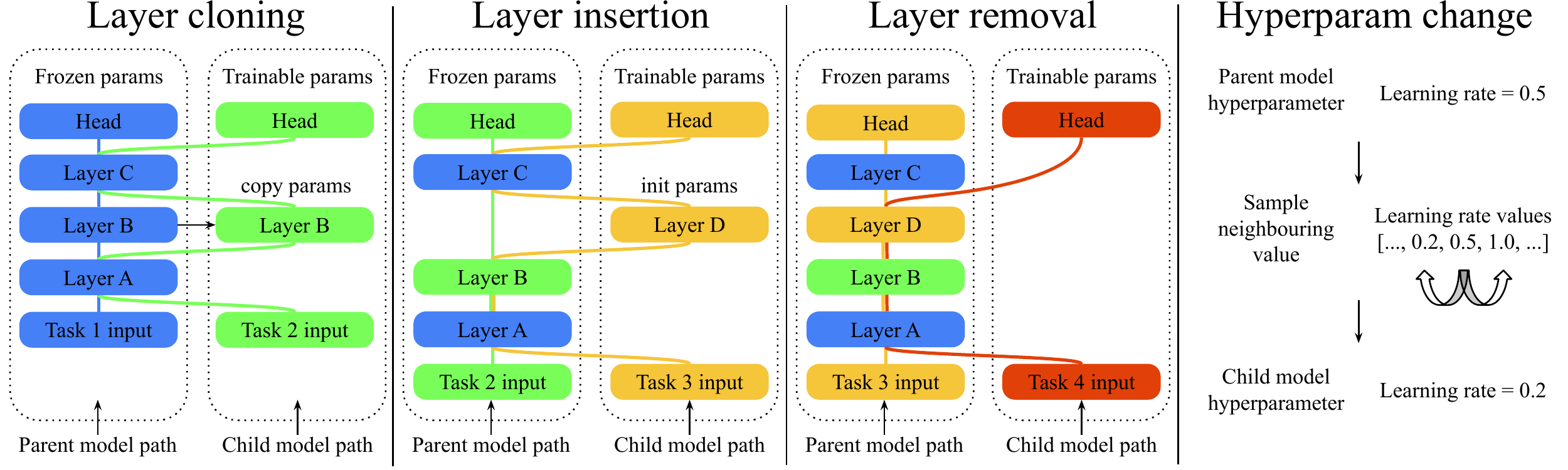}
\caption{
Graphical representation of the four types of mutations defined by the proposed method.
New models are generated by applying a subset of the possible mutations.}
\label{fig:mutations}
\end{figure}

\section{Method}
\label{section:method}
Deep neural networks are commonly defined as a sequence of layers that maps the input data into a prediction over the output space.
As a concrete example, we refer to the Visual Transformer (ViT) architecture \citep{Dosovitskiy2021AnII} that is used for the experimental phase.
ViT is composed of a sequence of different types of layers:
\begin{enumerate}  
\item \emph{Patch embedding}: the first layer of the model
maps the input image into a sequence of embedded tokens, each corresponding to a patch of the input image. 
\item \emph{Class token}: a classification token is prepended to the sequence. The final hidden state corresponding to this token is used as the aggregate sequence representation for classification tasks \citep{Devlin2019BERTPO}.
\item \emph{Position embedding}: the sequence representation is then augmented with 
an embedding that carries each patch positional information.
\item \emph{Transformer layers}: the sequence representation generated by the input layers is iteratively transformed by a stack of transformer layers \citep{Vaswani2017AttentionIA}.
\item \emph{Model head}: a final fully connected layer mapping the representation
produced by the top-most transformer layer for the class token
into the logits.
\end{enumerate}

\begin{figure}[t]
\centering
\hspace*{-8.4pt}
\begin{tikzpicture}
\node (image) at (0,0) {\includegraphics[width=1.0\linewidth]{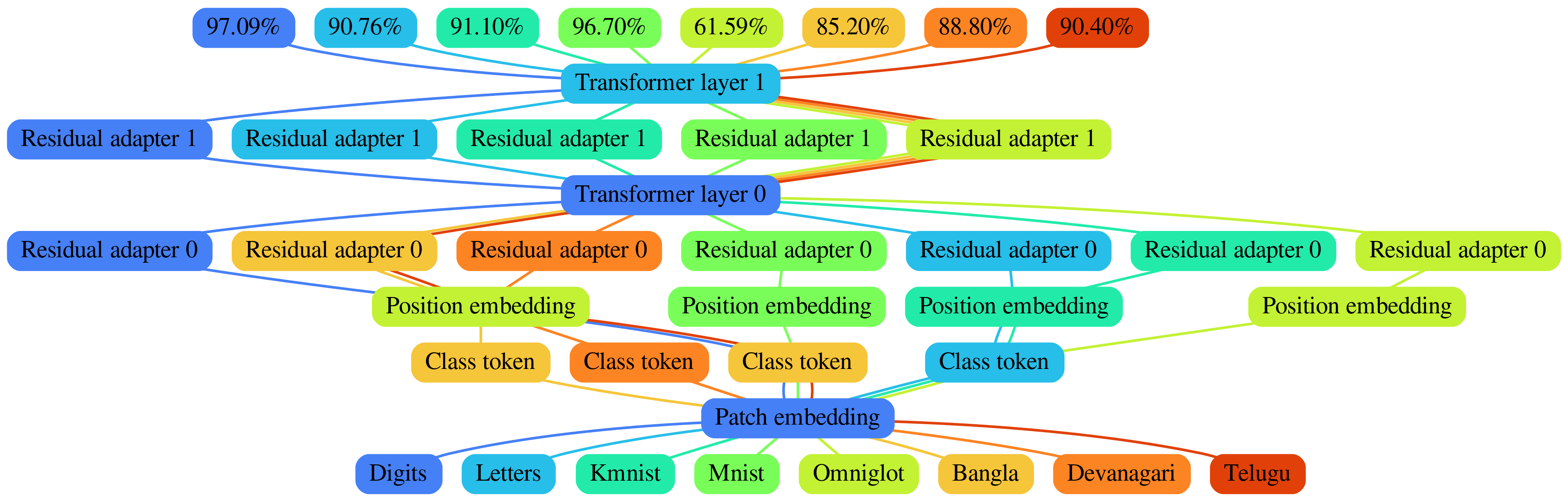}};
\node[stuff_fill] at (-1pt, 47.5pt) (TR1.0) {\tiny 2};
\node[stuff_fill] at (-145pt, 33.4pt) (RA1.0) {\tiny 3};
\node[stuff_fill] at (-88pt, 33.4pt) (RA1.1) {\tiny 3};
\node[stuff_fill] at (-31pt, 33.4pt) (RA1.2) {\tiny 3};
\node[stuff_fill] at (26pt, 33.4pt) (RA1.3) {\tiny 4};
\node[stuff_fill] at (83pt, 33.4pt) (RA1.4) {\tiny 3};
\node[stuff_fill] at (-1pt, 19.3pt) (TR0.0) {\tiny 1};
\node[stuff_fill] at (-145pt, 5.2pt) (RA0.0) {\tiny 4};
\node[stuff_fill] at (-88pt, 5.2pt) (RA0.1) {\tiny 4};
\node[stuff_fill] at (-31pt, 5.2pt) (RA0.2) {\tiny 5};
\node[stuff_fill] at (26pt, 5.2pt) (RA0.3) {\tiny 5};
\node[stuff_fill] at (83pt, 5.2pt) (RA0.4) {\tiny 3};
\node[stuff_fill] at (140pt, 5.2pt) (RA0.5) {\tiny 3};
\node[stuff_fill] at (197pt, 5.2pt) (RA0.6) {\tiny3};
\node[stuff_fill] at (-49.5pt, -8.9pt) (PO.0) {\tiny 2}; 
\node[stuff_fill] at (26pt, -8.9pt) (PO.1) {\tiny3};
\node[stuff_fill] at (86pt, -8.9pt) (PO.2) {\tiny 2};
\node[stuff_fill] at (173pt, -8.9pt) (PO3) {\tiny 2};
\node[stuff_fill] at (-59pt, -23pt) (CT.0) {\tiny 4};
\node[stuff_fill] at (-19pt, -23pt) (CT.1) {\tiny 5};
\node[stuff_fill] at (21pt, -23pt) (CT.2) {\tiny 4};
\node[stuff_fill] at (71pt, -23pt) (CT.3) {\tiny 2};
\node[stuff_fill] at (28pt, -37.1pt) (PA.0) {\tiny 1};
\end{tikzpicture}

\caption{Model graph representing the multitask network for the eight character classification tasks displayed on the bottom nodes,
generated by \href{https://youtu.be/Ld9gfmJT6Ig}{muNet with scale factor=0.3} (see Section~\ref{subsection:evo}).
This model was generated by the experiment repetition achieving the max average test accuracy across the eight tasks.
Each task is identified with a unique color.
Top nodes represent the head layer of each task, and display the validation accuracy for that task.
Each sequence of edges of the same color connecting a task input to its head, defines the layers sequence composing the model for each task.
Internal nodes are represented with the color of the task on which the parameters of the corresponding layer were trained last.
The number of unique tasks each layer have been trained on through the sequence of its ancestors is displayed in the top right corner label, {\small \textcircled{n}}.
}
\label{fig:graph30}
\end{figure}
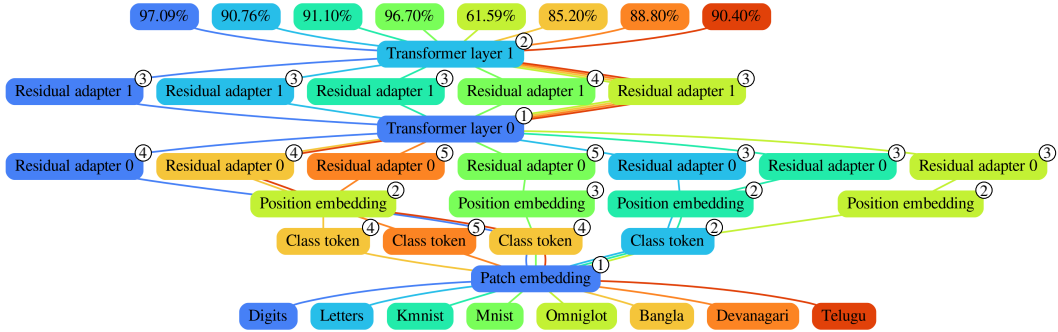

\subsection{Mutations}

We define four types of mutation (see Figure~\ref{fig:mutations}).
These mutations are designed to transform a pre-existing model, \emph{parent}, into a mutated version of it, \emph{child}.
The child model is defined by incremental 
changes to architecture, parameters and hyperparameters.
A child model may become a parent in following generations.
Each child model is trained 
on a single task,
but can leverage the knowledge accumulated by its ancestors on different tasks.
In the presented 
experiments (see Section~\ref{section:experiments}), a pretrained ViT model is used as the initial pre-existing model, \emph{root model}.
During the first evolutionary iteration, 
the root model is mutated by applying a subset of the possible mutations.
In subsequent iterations, ViT models that have already been subjected to mutations and training cycles can also be selected as a parent model.
The presented method instantiation 
allows the following mutation types:

\textbf{Layer cloning} \ \ Any layer of a parent model is by default shared with the child model in a frozen state,
so that the child models will not be able to apply gradient updates to the shared parameters, although they can flow gradients through the shared, frozen layers.
The layer cloning mutation allows a child model to create a trainable copy of any of the parent layers.
The head layer is always cloned since it always needs to be trainable.
If a child model is trained on a task different from the parent's task, then a new head layer is created with output shape matching the number of classes of the new task and zero initialized following \citet{Dosovitskiy2021AnII}.
Notice that the parent model parameters and architecture are immutable and cannot be affected by child mutations and subsequent training.
This is one of the features that provides strong guarantees against catastrophic forgetting.  

\textbf{Layer insertion} \ \ The model architecture can be mutated by inserting a new layer in between any two consecutive layers of the parent architecture.
In the instantiation of the method presented in this paper, we allow the insertion of residual adapter layers 
\citep{Rebuffi2017LearningMV,Houlsby2019ParameterEfficientTL}.
Residual adapters have been used with success as a parameter efficient method to adapt a pretrained model to a specific downstream task or domain.
We define residual adapters 
as a sequence of two fully connected layers with variable inner dimension size.
The Gelu non-linearity is applied on the inner representation \citep{Hendrycks2016GaussianEL}.
Layer normalization is applied to the input of the fully connected layers \citep{Ba2016LayerN}.
The second layer is zero initialized, to guarantee that its insertion does not alter the parent model representation at the start of the child training.

\textbf{Layer removal} \ \ This mutation removes a layer from the sequence that defines the parent model.
In the instantiation 
presented in this paper, layer removals are constrained to be applicable only to the top transformer layer.
This constraint avoids the knowledge and representation disruption that would result from removing internal layers, but still allows removing parameters and, combined with other types of mutations, to incrementally reach more parameter efficient configurations.

\textbf{Hyperparameter change} \ \ Every model is associated with a set of hyperparameters such as those of its optimizer, architecture and data preprocessing.
Any of the parent hyperparameters can be changed into a value sampled from the set of possible values defined for each hyperparameter.
In the instantiation of the method presented in this paper,
the sampling of numerical hyperparameters is constrained to the values that are neighbouring the parent value in the sorted list of possible values.
This constrains hyperparameters to be changed incrementally and biases the search toward the initial values of the root model, that in our application are the result of an extensive study aimed to identify a fine-tuning configuration for ViT models that is as generic as possible \citep{Steiner2021HowTT}.

Notice that, every hyperparameter of a child model 
is set to a single value.
However, considering that a child model can be interpreted as a continuation of its ancestors 
training with different hyperparameters, then the method can be regarded as capable of learning a schedule for each hyperparameter.
Furthermore, given that a different subset of layers is trainable for each ancestor, 
this approach can also be considered to be capable of learning a different optimizer schedule per layer. 

\subsection{Evolutionary algorithm}
\label{subsection:evo}
This section describes the novel evolutionary algorithm defined for the proposed method.
We refer to the first parent model,
used to initialize the evolutionary process,
as the \emph{root model}.
The root model can be either pretrained or randomly initialized.
During the evolutionary process,
the algorithm searches for an improved model
for a single task at a time, referred to as the \emph{active task}.
During the active phase of a task,
a population of models for the active task is evolved,
we refer to this as the \emph{active population}.
The active population is initialized with a set of \emph{seed parent models},
that includes the root model and the best model generated for each prior task.
Then, the active population is iteratively extended by:
\begin{enumerate} 
\item sampling a parent model from the active population,
\item producing a child model by applying to the parent model a sampled set of mutations,
\item performing cycles of training and evaluation 
to train and score the child model.
\end{enumerate}
Each model is assigned a \emph{score} that can be a function of multiple factors such as the validation quality or model cost metrics.
Early population pruning is performed by discarding the child models that did not achieve a better score than their parent.
At the end of each active phase for a task, only the model achieving best score is kept as part of the multitask model.
Tasks can become active multiple times, allowing for increased cross-tasks knowledge transfer.
Details of the evolutionary algorithm are provided below and in Algorithm~\ref{algo}.

\textbf{Parent sampling} \ \ Parent models are sampled among the models in the active population.
These candidate parent models are visited in order of score, starting with the highest scoring one. 
A candidate parent model, $m$, is selected as parent 
with probability:
\begin{equation}
p_{parent}(m) =\frac{1}{2}^{\#offsprings(m)}
\end{equation}
Where $\#offsprings(m)$ denotes the number of child models that have been generated so far for the active task by selecting model $m$ as parent. 
If the current candidate parent is rejected, then iteratively the model with the next best score is considered to be selected as parent with probability $p_{parent}(\cdot)$.
This approach can be interpreted as a back-off strategy following half-life exponential decay with $t_{\nicefrac{1}{2}}=1$.
This method prioritizes exploitation of high scoring models having few offsprings.
But also, in combination with early pruning, it automatically transitions toward a more exploratory behavior in case the higher scoring models are unable to generate improved offsprings.

\textbf{Mutations sampling} \ \ The mapping from a parent model into a child model is defined by a subset of the possible mutation actions.
The set of possible mutation actions includes:
a) one layer cloning action for each layer of the parent model,
b) one residual adapter insertion for each pair of consecutive transformer layers,
c) one top transformer layer removal action,
d) one hyperparameter change for each hyperparameter.
Each possible mutation is independently sampled for application with mutation probability, $\mu$.
For all the experiments reported in this paper, $\mu$ is set to 0.1.

\textbf{Child training} \ \ A newly sampled child model is trained on the active task for a given number of epochs. 
The model is evaluated on the validation set and scored after each epoch.
After training, only the parameters of the version of the child model achieving best score are retained.

\textbf{Scoring function} \ \ Each trained model is scored. 
The scoring function can be defined to optimize a mixture of factors such as quality, inference latency, training compute or model size depending on the application requirements and can change over time.
The experiments presented in this paper demonstrate the ability to control the quality/size trade-off by using the following scoring function:
\begin{equation}
score(m) = q(m) * s ^{\left( \frac{ \#accounted\mhyphen params(m) }{ \#root\mhyphen model\mhyphen params } \right)}
\end{equation}
Where $q(m)$ denotes the quality metric computed on the validation set.
$s\in\ ]0, 1]$ is the scale factor.
$\#root\mhyphen model\mhyphen params$ is the total number of parameters of the root model. 
And $\# accounted\mhyphen params(m)$ is the sum of parameters used by model $m$,
dividing each parameter count by the number of models 
sharing its use:
\begin{equation}
\#accounted\mhyphen params(m) = \sum_{p\in P(m)} \frac{1}{\#models(p)+1}
\end{equation}
Where $P(m)$ denotes the set of all parameters of $m$, and $\#models(p)$ is the count of models for tasks different from the active task that are currently using this parameter. 
The scaling factor, $s$, allows to control the size of the generated multitask model, and achieve different quality/size trade-offs.

Note that, the defined evolutionary algorithm guarantees that once a model has been trained, its architecture and the parameters storing its knowledge and cannot be altered.
Nonetheless, new models can access its knowledge or even extend it to improve it or specialize it.
Therefore, this method provides immunity against common problems of multitask models:
    1) \emph{catastrophic forgetting}, since the knowledge of a trained model is always preserved,
    2) \emph{negative transfer}, since the method 
    automates the selection of the knowledge most relevant for each new task,
    3) \emph{gradients interference}, since within each training cycle each parameter can receive gradients only from one source.

\section{Experiments} 
\label{section:experiments}
This section describes the experiments conducted to analyze the properties of the proposed method and test their generality. The proposed method is referred to as “multitask network” or for brevity muNet.
All the experiments reported are reproducible by using: 1)~The ViT checkpoints 
and model definition library published by \citet{Steiner2021HowTT}, 2)~the published code of the proposed method, 
3)~datasets publicly available via the \href{https://www.tensorflow.org/datasets/catalog/overview}{Tensorflow Datasets image classification catalog}.
All the experiments are executed on a TPUv3 machine with 8 cores \citep{Jouppi2017IndatacenterPA}.

The default ViT configuration used, is the one identified by \citet{Steiner2021HowTT} as the most generic and best performing for ViT fine-tuning: SGD optimizer with 0.9 momentum and 0.01 learning rate, using cosine decay schedule with 10\% warm up, 512 batch size, no weight decay, and gradient clipping at global norm 1.
During auto-tuning experiments, the evolutionary algorithm can change the following hyperparameters of the optimizer, image preprocessing and architecture:
\begin{itemize}
\item  learning rate $\in$ [0.0001, 0.0002, 0.0005, 0.001, 0.002, 0.005, \textbf{0.01}, 0.02, 0.05, 0.1, 0.2, 0.5]
\item learning rate schedule $\in$ [constant, \textbf{cosine}, restarts]
\item learning rate schedule warm up ratio $\in$ [0.01, 0.02, 0.05, \textbf{0.1}, 0.2, 0.3, 0.4]
\item momentum $\in$ [0.7, 0.8, 0.85, \textbf{0.9}, 0.95, 0.98, 0.99]
\item nesterov update $\in$ [\textbf{False}, True]
\item cropped area range min $\in$ [0.01, 0.02, \textbf{0.05}, 0.1, 0.2, 0.5, 1.0]
\item cropped aspect ratio range min $\in$ [0.25, 0.5, \textbf{0.75}, 1.0]
\item flip left/right $\in$ [False, \textbf{True}]
\item brightness delta $\in$ [\textbf{0.0}, 0.01, 0.02, 0.05, 0.1, 0.2]
\item contrast delta $\in$ [\textbf{0.0}, 0.01, 0.02, 0.05, 0.1, 0.2]
\item saturation delta $\in$ [\textbf{0.0}, 0.01, 0.02, 0.05, 0.1, 0.2]
\item hue delta $\in$ [\textbf{0.0}, 0.01, 0.02, 0.05, 0.1, 0.2]
\item image size $\in$ [multiples of root model's patch size]
\item residual adapters inner dimension $\in$ [8, 16, \textbf{32}, 64, 128]
\end{itemize}

Bold values are the defaults.
This search space is a parametrization of the ViT model
configuration as defined by the published ViT model library,
except for the "residual adapters inner dimension",
which has been added 
for the layer insertion mutation introduced by our evolutionary algorithm.

\begin{figure}[t]
\centering
\begin{minipage}{.5\textwidth}
  \centering
  \includegraphics[width=1.0\linewidth]{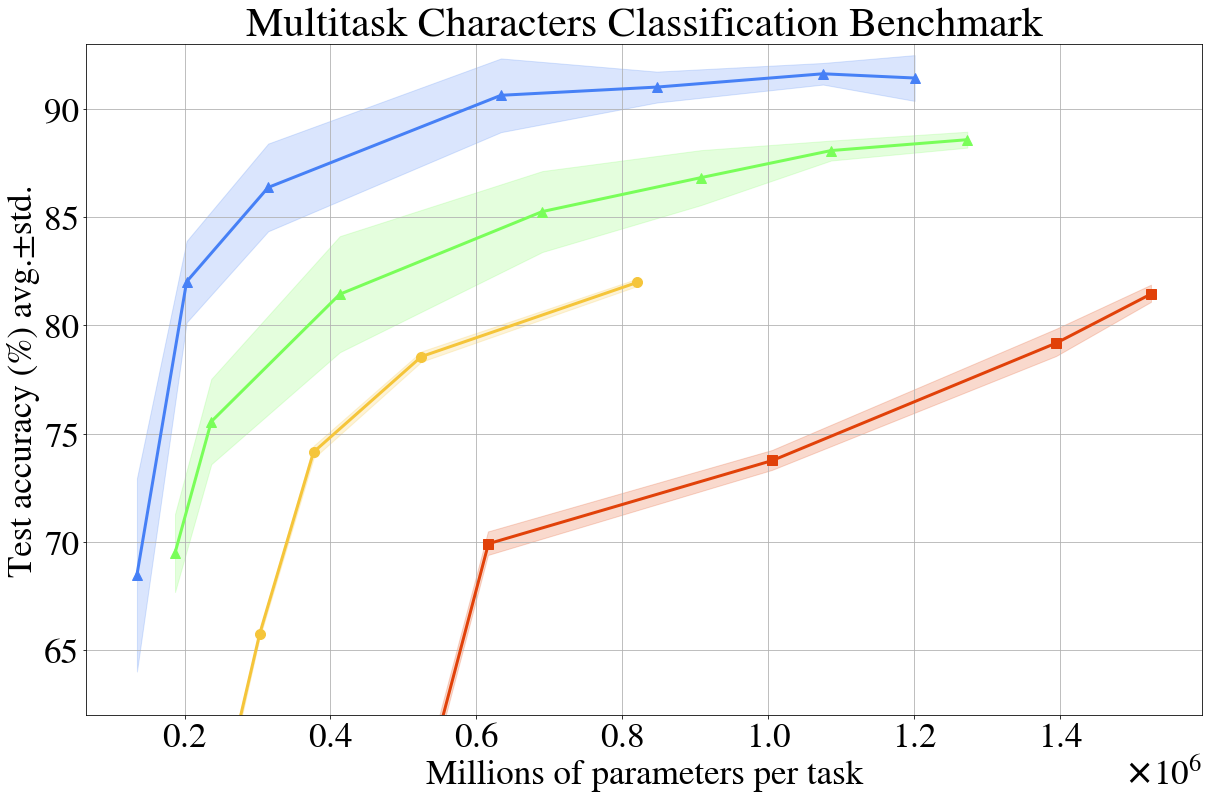}
\end{minipage}%
\begin{minipage}{.5\textwidth}
  \centering
  \includegraphics[width=1.0\linewidth]{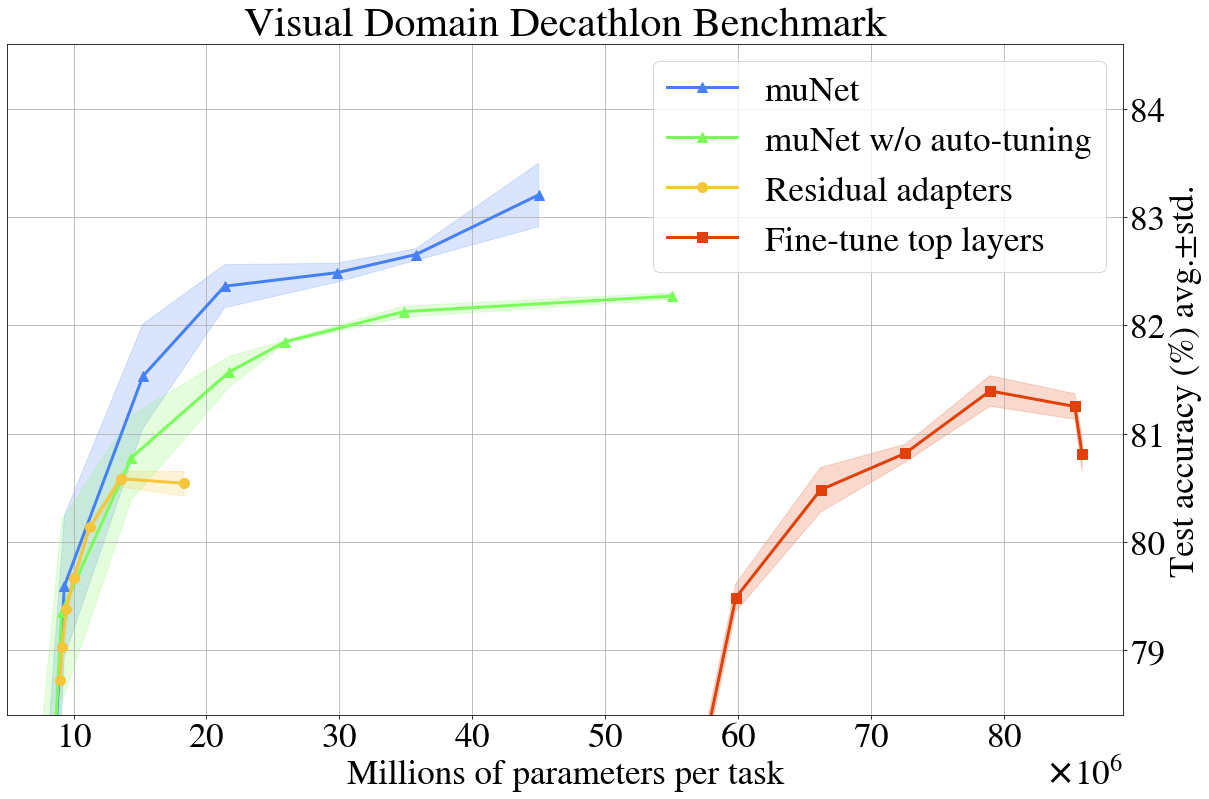}
\end{minipage}
\caption{Comparison between standard fine-tuning techniques and the proposed method, with and without hyperparameters auto-tuning enabled, on the Multitask Character Classification Benchmark (left: see Section~\ref{subsection:chars}) and the Visual Domain Decathlon Benchmark (right: see Section~\ref{subsection:deca}).}
\label{fig:curves}
\end{figure}

\subsection{Multitask Character Classification Benchmark}
\label{subsection:chars}

The first benchmark 
is designed to ease reproducibility and enable fast development iterations.
It is composed of 8 publicly available character classification tasks (see Table~\ref{table:metadata}).
For each 
we identify non-overlapping training, validation and test splits (see Table~\ref{table:splits}).

The root model for this benchmark, is a ViT Ti/16 pretrained on the imagenet-21k dataset with weight decay 0.1, no stochastic depth, no dropout and “light1” augmentation as defined in \citet{Steiner2021HowTT}.
As the intent of this benchmark is to allow for fast iterations, 
the root model architecture is capped to use only 3 of the 12 transformer layers provided by the referred configuration. 
The default image size for this benchmark is set to $32{\times}32$ pixels, since it is a multiple of the patch size (16)  
that is close to the resolution in which images are provided for most of the tasks in this benchmark (28).

Baseline models are initialized with the same parameters of the root model 
and are fine-tuned for a total of 80 
epochs on each task.
8 model replicas are trained in parallel, one on each core.
The version of the baseline model 
achieving the best validation quality among all the periodic evaluations performed by all 
replicas is evaluated on the 
test set.
%
The muNet experiments are allocated an equivalent 
budget 
of training steps. 
Each task is given 2 active task iterations, and each such iteration is composed of 8 child model generation phases.
During each generation phase, 8 child models are sampled and trained in parallel, one on each of the 8 TPUv3 cores.
Each child model is trained for 5 epochs,
in order to achieve a training budget equivalent to that allocated for the baseline model:
\begin{equation}
\label{equation:epochs}
\# baseline\mhyphen epochs = \# muNet\mhyphen epochs * \#generations * \#task\mhyphen iterations
\end{equation}
To smooth the distribution of compute over the set of tasks, the amount of the training performed between validations is capped to: $min(1\ epoch, 100\ batches)$.

\textbf{Results} \ \ 
The experiments demonstrate the ability of the proposed method to achieve better quality/size trade-offs compared to standard fine-tuning techniques.
Metrics are summarized in Table~\ref{table:chars} and \ref{table:chars-extra}.
Figure~\ref{fig:curves} (left) displays a graphical summary of the 
trade-offs achieved 
by the different methods.
The horizontal axis measures the average number of parameters per task.
The vertical axis measures the mean accuracy achieved on the final test sets of the 8 tasks.
Each experiment was repeated 5 times for each model configuration using different random seed values for each repetition.  
The vertical coordinate of the plotted curves represents the average quality and the shaded area represents the standard deviation computed across the experiment repetitions.

The rightmost point of the “Fine-tune top layers” curve represents quality/size achieved by fully fine-tuning a distinct 
copy of the root model for each each of the 8 tasks. 
The horizontal coordinate of full fine-tuning matches the size of the ViT Ti/16 3 layers model used as root model: \textasciitilde1.48M parameters.
The next point following the “Fine-tune top layers” curve from right to left, represents the model configuration having all the layers before the first transformer layer frozen and shared across the 8 tasks.
This configuration achieves a lower quality with fewer parameters per task compared to full fine-tuning.
The same trend continues as more transformer layers are frozen and shared, until the last configuration matches the multi-head architecture, where all the layers are frozen and shared except for each individual task head.

The rightmost point of the “Residual adapters” curve represents the quality/size achieved by the architecture configuration that shares all the parameters of the root model, and only head and residual adapters with hidden dimension of 512 are trainable.
Following the “Residual adapters” curve right to left, the hidden dimension of each residual adapter is halved at each step, resulting in a monotonic decrease of both quality and model size.
Different trade-offs for the proposed method are are achieved by using different scale factors: $s\in$ \{0.02, 0.3, 0.7, 0.9, 0.95, 0.98, 1\}.
The rightmost point in the curve uses scale factor of 1, resulting in no size penalty: $score(m) = q(m)$.
We also compare against a version of the proposed method with ablated hyperparameters auto-tuning.

\begin{table}[t]
\caption{
Models comparison on the Multitask Character Classification Benchmark.
Quality is measured as the test accuracy averaged across the 8 tasks.
The table reports the max, average and standard deviation of the quality achieved by 5 experiment repetitions.
Model size is measured as the average number of parameters per tasks.
Compute is measured as TPUv3 core-hours.
Videos displays the evolution of the best experiment repetition.
Tasks details are reported in Table~\ref{table:chars-extra}.
}
\label{table:chars}
\centering
\begin{tabular}{lccccc}
\toprule
 & \multicolumn{2}{c}{Test acc. (\%)} & Params/task & Compute & Video\\
 \cmidrule(r){2-3}
Model &    Max &   Avg.$\pm$Std. & \multicolumn{1}{c}{($\times10^6$)} & (core-h) & (youtu.be/... )
\\
\midrule
 Multi-head &  20.79 &  20.25$\pm$0.65 &  0.23 
 & 4.32 &
\href{https://youtu.be/LIvOCmF1aRk}{\small{\texttt{LIvOCmF1aRk}}}\\

 Full fine-tuning &  82.20 &  81.46$\pm$0.40 &  1.53 
 & 4.91 &
\href{https://youtu.be/_fikzbxS_ZY}{\small{\texttt{\_fikzbxS\_ZY}}}\\

 Residual adapters dim=512 &  82.28 &  81.98$\pm$0.17 &  0.82 
 & 5.01 &
\href{https://youtu.be/R3ETGxo9CWE}{\small{\texttt{R3ETGxo9CWE}}}\\

 muNet w/o auto-tuning &  88.96 &  88.57$\pm$0.37 &  1.27 
 & 4.81 &
 \href{https://youtu.be/CQdeP1mpr-8}{\small{\texttt{CQdeP1mpr-8}}}\\

 muNet scale factor=0.3 &  84.94 &  82.01$\pm$1.88 &  0.20 
 & 4.63 &
 \href{https://youtu.be/Ld9gfmJT6Ig}{\small{\texttt{Ld9gfmJT6Ig}}}\\

 muNet &  92.98 &  91.41$\pm$1.06 &  1.20 
 & 5.02 &
\href{https://youtu.be/-xOl3lJV4fw}{\small{\texttt{-xOl3lJV4fw}}}\\

\bottomrule
\end{tabular}
\end{table}

The proposed method achieves the best quality across the spectrum of model sizes.
The best muNet model 
improves the best quality achieved by full fine-tuning by 13\%, while using 21\% fewer parameters per task.
muNet with scale factor 0.3 outperforms 
all the baseline models of all types in both quality and size dimensions.
The configuration with scale factor 0.3 outperforms the quality of the best full fine-tuning model,
while 
using less parameter than the multi-head model.
The smaller size is achieved because the evolutionary search 
converged to use only 2 of the 3 transformer layers 
for all tasks.
Figure~\ref{fig:graph30} displays the multitask architecture generated by the best experiment repetition with scale factor 0.3.
The layers with most parameters are shared across all tasks: transformer layers (444,864 parameters) and patch embedding (147,648).
While the smaller layers are branched for specialization: class token (192), position embedding (960), residual adapters (12,896).
Comparing with muNet without auto-tuning allows us to assess 
the 
impact of the 
hyperparameter evolution.

Furthermore, residual adapters ablation experiments are then performed to analyze the effects of the layer insertion mutation 
(see Figure~\ref{fig:ablate-ra}).
For scale factors above 0.9, it is possible to achieve equivalent performance within noise even without residual adapters. 
However, for lower scale factors, residual adapters seem critical to achieve significantly better quality/size trade-offs.
Nevertheless, muNet with neither residual adapters nor auto-tuning still outperforms the residual adapters baseline.

\subsection{Visual Domain Decathlon Benchmark}
\label{subsection:deca}

The generality of the proposed method is tested with more challenging tasks, longer training, and bigger models.
The Visual Domain Decathlon Benchmark \citep{hakanbilensylvestrerebuffitomasjakab2017} consists of 10 image classification tasks
that have been explicitly
selected to represent different domains (see Table~\ref{table:metadata}),
thus providing a more challenging context for knowledge transfer.
All datasets consist of images with $72{\times}72$ resolution.
Thus, the default resolution is set to the next multiple of the patch size: $80{\times}80$.

While the experiments on the Multitask Character Classification Benchmark provided comparison in a fast training setup, this benchmark is configured for longer and more expensive training.
Thus, baseline models are trained for a total of 480 epochs on each task.
The proposed method is again provided with an equivalent training budget.
Two iterations are performed over the task set.
For every iteration on each active task, 8 batches of child models are generated.
That, following equation~\ref{equation:epochs}, results in 30 epochs of training per child model.
Epochs are again capped to 200 batches.
As a root model, we use a ViT B/16 pretrained on the imagenet-21k dataset with weight decay 0.1, no stochastic depth, no dropout and “medium1” augmentation as defined in \citet{Steiner2021HowTT}.
All ViT B/16 12 transformer layers are used, for a total of 85.6M parameters.
This is \textasciitilde60 times bigger than the root model used for the experiments on the Multitask Character Classification Benchmark.
The batch size is halved to 256, to fit the memory requirements. 
The number of experiment repetitions is decreased from 5
to 3 due to the higher experiment cost.
Other configuration details are unchanged.

\begin{table}[t]
\caption{
Models comparison on the Visual Domain Decathlon Benchmark.
Quality is measured as the test accuracy averaged across the 10 tasks.
The table reports the max, average and standard deviation of the quality achieved by 3 experiment repetitions. 
Model size is measured as the average number of parameters per tasks.
Compute is measured as TPUv3 core-hours.
Videos displays the evolution of the best repetition.
Tasks details are reported in Table~\ref{table:deca-extra}.
}
\label{table:deca}
\centering
\begin{tabular}{lcccccc}
\toprule
 & \multicolumn{2}{c}{Test acc. \%} & Params/task & Compute & Video\\
  \cmidrule(r){2-3}
Model & max & avg.$\pm$std. & ($\times10^6$) & (core-h) & (youtu.be/...) \\
\midrule
 Multi-head &  49.77 &  49.73$\pm$0.03 &  8.81 
 & 176.34 &
 \href{https://youtu.be/7UgPZYgh53U}{\small{\texttt{7UgPZYgh53U}}}\\

 Unfreeze above 1st tr. layer &  81.51 &  81.39$\pm$0.14 &  78.98 
 & 183.67 &
\href{https://youtu.be/vNo-j150nA0}{\small{\texttt{vNo-j150nA0}}}\\

 Full fine-tuning &  81.03 &  80.81$\pm$0.15 &  85.91 
 & 185.34 &
\href{https://youtu.be/BK7AW95ii4s}{\small{\texttt{BK7AW95ii4s}}}\\

 Residual adapters dim=512 &  80.64 &  80.54$\pm$0.11 &  18.28  
 & 180.86 &
 \href{https://youtu.be/oWiniz6F2Lw}{\small{\texttt{oWiniz6F2Lw}}}\\

 muNet w/o auto-tuning &  82.31 &  82.27$\pm$0.03 &  55.04 
 & 181.05 &
\href{https://youtu.be/P0SBFOuyj0s}{\small{\texttt{P0SBFOuyj0s}}}\\

 muNet scale factor=0.3 &  80.11 &  79.59$\pm$0.65 &  9.30 
 & 165.34 &
\href{https://youtu.be/THyc5lUC_-w}{\small{\texttt{THyc5lUC\_-w}}}\\

 muNet &  83.58 &  83.20$\pm$0.29 &  45.00 
 & 185.45 &
 \href{https://youtu.be/2scExBaHweY}{\small{\texttt{2scExBaHweY}}}\\

\bottomrule
\end{tabular}
\end{table}

\textbf{Results} \ \ 
Figure~\ref{fig:curves} presents a graphical summary of the comparison between the different methods,
while Table~\ref{table:deca} 
reports a numerical summary.
Again, we observe that the proposed method outperforms standard fine-tuning methods.
Once more, auto-tuning provides a significant contribution to improvements in both quality and size dimensions.
The evolution of the best "scale factor=0.3" shows that most of the layers specialization happens in the smaller layers and 1 to 2 transformer layers are dropped for each task (see Figure \ref{fig:deca-30}).
Differently from the previous results,
we observe that the best fine-tuning performance is obtained by fine-tuning only the layers 
above the first transformer layer.

Individual task accuracies (see Table~\ref{table:deca-extra}) show that muNet models archives significant gains on the smaller tasks,
this can be expected as the smaller tasks have fewer data to train on and can benefit the most from knowledge transfer.
But also, significant gains can be observed on larger tasks like imagenet or cifar100.
Baseline models validation curves (see Figure~\ref{fig:crvs}) show lower variance.

Analyzing the knowledge exchange dynamics (see Figures~\ref{fig:flow-deca} and \ref{fig:flow-char}), we notice that specialized tasks with small datasets, such as vgg-flowers, aircraft and ucf101, reuse knowledge from multiple tasks, but no other task reuses the parameters fine-tuned by them.
Inversely, generic tasks with large datasets, such as imagenet12, contribute to all other tasks, but their 
model results mostly trained on their 
data.

The distributions of the hyperparameters sampled by the auto-tuning algorithm (see Figure~\ref{fig:hps}) show that the distributions shift in accordance with the reward system.
For example, as the size penalty increases, distributions shift toward sampling smaller adapters and less transformer layers.

\section{Related work}

The success of \textbf{transfer-learning} applications hinge on adequate prior knowledge selection, that avoids common negative-transfer pitfalls \citep{Rosenstein2005ToTO,Wang2019CharacterizingAA}.
Common solutions rely on data or model selection techniques
\citep{Zhang2020ASO,Mensink2021FactorsOI}.
Models trained jointly on \textbf{multiple tasks} 
can be affected by negative gradients interference when parameters receive gradients from multiple sources \citep{Chen2018GradNormGN,Yu2020GradientSF}, and by catastrophic forgetting of prior knowledge as new tasks are learned.
Catastrophic forgetting is also critical for \textbf{continual learning} or life long learning applications
\citep{McCloskey1989CatastrophicII,French1999CatastrophicFI}.
These knowledge loss problems can be alleviated with weighted combination of tasks \citep{Liu2019LossBalancedTW,Sun2020ERNIE2A} and gradient transformation methods \citep{Chen2018GradNormGN,Sener2018MultiTaskLA,Kendall2018MultitaskLU}.
Stronger guarantees are provided by methods that compartmentalize task specific knowledge in dedicated parameter subsets \citep{Rebuffi2017LearningMV,Houlsby2019ParameterEfficientTL,Rusu2016ProgressiveNN,Rosenfeld2020IncrementalLT}.

The automation of \textbf{hyperparameter tuning} has been commonly addressed with Bayesian optimization \citep{Srinivas2010GaussianPO,Bergstra2011AlgorithmsFH,Snoek2012PracticalBO},
evolutionary methods have also been explored 
\citep{Jaderberg2017PopulationBT,Zhang2011EvolutionaryCM}.
Hyperparameter tuning can be considered related to the \textbf{neural architecture search} (NAS), as architectures can be defined by the selection 
of architectural hyperparameters.
Initially, NAS methods have been based on reinforcement learning 
\citep{Zoph2017NeuralAS,Tan2019MnasNetPN}.
Sample efficient evolutionary approaches have been also proposed \citep{Real2019RegularizedEF,Maziarz2018EvolutionaryNeuralHA},
Furthermore, more efficient parameter-sharing approaches have been proposed \citep{Pham2018EfficientNA,Liu2019DARTSDA,Kokiopoulou2019FastTA} that connect the NAS field with the one of \textbf{routing networks}  \citep{Fernando2017PathNetEC,Maziarz2019GumbelMatrixRF}.

\section{Conclusion}
\label{section:conclusions}

We introduced a novel method that can evolve pretrained 
ML models into multitask systems capable 
of jointly solving many tasks.
Empirical evidences show that the method can achieve improved quality and efficiency compared to common fine-tuning techniques.
We also presented a novel evolutionary approach that is employed for both identifying the knowledge most suitable for any new task and also dynamically tuning the hyperparameters of the system components. 
Furthermore, the presented evolutionary method can automatically adjust the exploration/exploitation balance without requiring to tune any additional meta-parameter.
The generated multitask systems are immune from catastrophic forgetting, negative transfer and gradients interference.

Overall, the approaches demonstrated in this work are encouraging signs that a much-more-automated, incrementally extensible machine learning system for handling thousands or millions of tasks is achievable.
We show that starting with a high quality baseline model and combining this with a novel evolutionary search procedure can efficiently tackle new tasks and create a single multi-task model with high accuracy across all tasks in a completely automated manner.
Future work can continue to 
build toward a system that can handle thousands and then millions of tasks, across multiple modalities.

\textbf{Limitations} \ \ To contain results variance, it is important to control the ratio between the size of the search space and the available exploratory budget.
This can be challenging considered that any incremental addition to the search space causes an exponential increase in the number of possible model configurations. For example, adding one clonable layer doubles the number of possible configurations, and adding a new hyperparameter with 10 possible values increases it by a decimal order of magnitude.
While, an increase in compute 
leads to a linear increase in exploration budget.
Method details such as incremental hyperparameter mutation and exponential decay sampling,
contribute to control the variance by implicitly imposing a prior distribution over the search space.

\textbf{Societal impact} \ \ 
Improvements in the efficiency, ease-of-use and automation of contemporary ML methodologies such as those proposed in this work, can broaden the accessibility of ML approaches to a much wider audience, as less ML-specific knowledge is needed to achieve a high-quality 
model for a new task.
This approach can 
contribute to reducing the energy usage and carbon footprint of ML applications.
Granting access to the broader community to novel techniques and systems allows to democratize the ground-breaking advancements in automation and achievements of new capabilities.



\medskip
{
\small
\bibliographystyle{abbrvnat}
\bibliography{neurips_2022}
}

\clearpage

\appendix

\section{Assets details}
\label{app:assets} 
This section reports details of the datasets, code and model checkpoints used for the presented empirical study.
The code implementing the proposed method is published at \href{https://github.com/google-research/google-research/tree/master/muNet}{https://github.com/google-research/google-research/tree/master/muNet} and distributed under the \href{https://www.apache.org/licenses/LICENSE-2.0}{Apache License 2.0}.
The ViT model definition and checkpoints published by \citet{Steiner2021HowTT} are available at  \href{https://github.com/google-research/vision_transformer}{https://github.com/google-research/vision\_transformer} and distributed under the \href{https://github.com/google-research/vision_transformer/blob/main/LICENSE}{Apache License 2.0}.
These resources allow to reproduce all the reported experiments.
Tables \ref{table:metadata}, \ref{table:splits} and \ref{table:ds-refs} report details of the datasets used.

\vspace{100pt}
\begin{table}[h]
  \caption{
  Datasets metadata:
  1) name,
  2) short description,
  3) number of classes,
  4) input image resolution,
  5) number of train samples,
  6) number of validation samples,
  7) number of test samples.
  }
  \label{table:metadata}
  \centering
  \begin{tabular}{lcccccc}
    \toprule
    Name     & Description & Cls. & Res.  & Train & Valid. & Test \\
    \midrule
    
\multicolumn{7}{c}{\textbf{Multitask Character Classification Benchmark}} \\

\href{https://www.tensorflow.org/datasets/catalog/emnist#emnistdigits}{emnist/digits} & classify digits & 10 & 28 & 228k & 12k & 40k
\\
\href{https://www.tensorflow.org/datasets/catalog/emnist#emnistletters}{emnist/letters}
& classify letters & 37 & 28 & 84.4k
 & 4.4k & 14..8k
\\
\href{https://www.tensorflow.org/datasets/catalog/kmnist}{kmnist}
& classify Japanese chars. & 10 & 28
& 57k & 3k & 10k
\\
\href{https://www.tensorflow.org/datasets/catalog/mnist}{mnist}
& classify digits & 10 & 28
& 57k & 3k & 10k
\\
\href{https://www.tensorflow.org/datasets/catalog/omniglot}{omniglot}
& cls. 50 alphabets chars. & 1623 & 105
& 19.3k & 2.7k & 3.1k
\\
\href{https://www.tensorflow.org/datasets/catalog/cmaterdb#cmaterdbbangla_default_config}{cmaterdb/bangla}
& cls. Bangla numerals
& 10 & 32
& 4.75k & 250 & 1k
\\
\href{https://www.tensorflow.org/datasets/catalog/cmaterdb#cmaterdbdevanagari}{cmaterdb/devanagari}
& cls. Devangari numerals
& 10 & 32
& 2.38k & 125 & 500
\\
\href{https://www.tensorflow.org/datasets/catalog/cmaterdb#cmaterdbtelugu}{cmaterdb/telugu}
& cls. Telugu numerals
& 10 & 32
& 2.38k & 125 & 500
\\
 
\midrule
\multicolumn{7}{c}{\textbf{Visual Domain Decathlon Benchmark}} \\

\href{https://www.tensorflow.org/datasets/catalog/visual_domain_decathlon#visual_domain_decathlonaircraft_default_config}{aircraft}
&  classify aircrafts
& 100 & 72 
& 3.33k & 1.67k & 1.67k
\\
\href{https://www.tensorflow.org/datasets/catalog/visual_domain_decathlon#visual_domain_decathloncifar100}{cifar100}
& classify image subject
& 100 & 72 
& 50k & 5k & 5k
\\
\href{https://www.tensorflow.org/datasets/catalog/visual_domain_decathlon#visual_domain_decathlondaimlerpedcls}{daimlerpedcls}
& classify pedestrians
& 2 & 72 
& 23.5k & 2.9k & 2.9k
\\
\href{https://www.tensorflow.org/datasets/catalog/visual_domain_decathlon#visual_domain_decathlondtd}{dtd}
& classify textures
& 47 & 72 
& 1.9k & 940 & 940
\\
\href{https://www.tensorflow.org/datasets/catalog/visual_domain_decathlon#visual_domain_decathlongtsrb}{gtsrb}
& cls. German traffic signs
& 43 & 72 
& 31.4k & 3.9k & 3.9k
\\
\href{https://www.tensorflow.org/datasets/catalog/visual_domain_decathlon#visual_domain_decathlonimagenet12}{imagenet12}
& cls. annotated concept
& 1000 & 72 
& 1.23M & 24.5k & 24.5k
\\
\href{https://www.tensorflow.org/datasets/catalog/visual_domain_decathlon#visual_domain_decathlonomniglot}{omniglot}
& cls. 50 alphabets chars.
& 1623 & 72 
& 17.8k & 3.2k & 3.2k
\\
\href{https://www.tensorflow.org/datasets/catalog/visual_domain_decathlon#visual_domain_decathlonsvhn}{svhn}
& classify house numbers
& 10 & 72 
& 47.2k & 13k & 13k
\\
\href{https://www.tensorflow.org/datasets/catalog/visual_domain_decathlon#visual_domain_decathlonucf101}{ucf101}
& cls. dynamic images
& 101 & 72 
& 7.6k & 976 & 976
\\
\href{https://www.tensorflow.org/datasets/catalog/visual_domain_decathlon#visual_domain_decathlonvgg-flowers}{vgg-flowers}
& classify flowers
& 102 & 72 
& 1k & 510 & 510
\\
    \bottomrule
  \end{tabular}
\end{table}

\begin{table}[h]
  \caption{
  Datasets splits configuration.
  Datasets names match with the Tensorflow Datasets Catalogs identification strings and link to the corresponding catalog page.
  Datasets splits are represented with the \href{https://www.tensorflow.org/datasets/splits}{standard Tensorflow Datasets format}.
  }
  \label{table:splits}
  \centering
  \begin{tabular}{lccc}
    \toprule
    &\multicolumn{3}{c}{Splits}                   \\
    \cmidrule(r){2-4}
    Name     & Train & Validation & Test \\
    \midrule
    
\multicolumn{4}{c}{\textbf{Multitask Character Classification Benchmark}} \\

\href{https://www.tensorflow.org/datasets/catalog/emnist#emnistdigits}{emnist/digits} &
train[5\%:] & train[:5\%] & test 
\\
\href{https://www.tensorflow.org/datasets/catalog/emnist#emnistletters}{emnist/letters} &
train[5\%:] & train[:5\%] & test
\\
\href{https://www.tensorflow.org/datasets/catalog/kmnist}{kmnist} &
train[5\%:] & train[:5\%] & test
\\
\href{https://www.tensorflow.org/datasets/catalog/mnist}{mnist} &
train[5\%:] & train[:5\%] & test
\\
\href{https://www.tensorflow.org/datasets/catalog/omniglot}{omniglot} &
train & small1 & small2
\\
\href{https://www.tensorflow.org/datasets/catalog/cmaterdb#cmaterdbbangla_default_config}{cmaterdb/bangla} &
train[5\%:] & train[:5\%] & test
\\
\href{https://www.tensorflow.org/datasets/catalog/cmaterdb#cmaterdbdevanagari}{cmaterdb/devanagari} &
train[5\%:] & train[:5\%] & test
 \\
\href{https://www.tensorflow.org/datasets/catalog/cmaterdb#cmaterdbtelugu}{cmaterdb/telugu} &
train[5\%:] & train[:5\%] & test
 \\
 
\midrule
\multicolumn{4}{c}{\textbf{Visual Domain Decathlon Benchmark}} \\

\href{https://www.tensorflow.org/datasets/catalog/visual_domain_decathlon#visual_domain_decathlonaircraft_default_config}{aircraft} &
train & validation[:50\%] & validation[50\%:]
\\
\href{https://www.tensorflow.org/datasets/catalog/visual_domain_decathlon#visual_domain_decathloncifar100}{cifar100} &
train & validation[:50\%] & validation[50\%:]
\\
\href{https://www.tensorflow.org/datasets/catalog/visual_domain_decathlon#visual_domain_decathlondaimlerpedcls}{daimlerpedcls} &
train & validation[:50\%] & validation[50\%:]
\\
\href{https://www.tensorflow.org/datasets/catalog/visual_domain_decathlon#visual_domain_decathlondtd}{dtd} &
train & validation[:50\%] & validation[50\%:]
\\
\href{https://www.tensorflow.org/datasets/catalog/visual_domain_decathlon#visual_domain_decathlongtsrb}{gtsrb} &
train & validation[:50\%] & validation[50\%:]
\\
\href{https://www.tensorflow.org/datasets/catalog/visual_domain_decathlon#visual_domain_decathlonimagenet12}{imagenet12} &
train & validation[:50\%] & validation[50\%:]
\\
\href{https://www.tensorflow.org/datasets/catalog/visual_domain_decathlon#visual_domain_decathlonomniglot}{omniglot} &
train & validation[:50\%] & validation[50\%:]
\\
\href{https://www.tensorflow.org/datasets/catalog/visual_domain_decathlon#visual_domain_decathlonsvhn}{svhn} &
train & validation[:50\%] & validation[50\%:]
\\
\href{https://www.tensorflow.org/datasets/catalog/visual_domain_decathlon#visual_domain_decathlonucf101}{ucf101} &
train & validation[:50\%] & validation[50\%:]
\\
\href{https://www.tensorflow.org/datasets/catalog/visual_domain_decathlon#visual_domain_decathlonvgg-flowers}{vgg-flowers} &
train & validation[:50\%] & validation[50\%:]
\\
    \bottomrule
  \end{tabular}
\end{table}

\begin{table}[h]
  \caption{
  Datasets reference and license.
  }
  \label{table:ds-refs}
  \centering
  \begin{tabular}{lll}
    \toprule
    Name     & Reference & License \\
    \midrule
    
\multicolumn{3}{c}{\textbf{Multitask Character Classification Benchmark}} \\

\href{https://www.tensorflow.org/datasets/catalog/emnist#emnistdigits}{emnist/digits} &
\citep{Cohen2017EMNISTEM} & 
\href{https://creativecommons.org/licenses/by/4.0/}{Creative Commons Attribution 4.0 License}
\\
\href{https://www.tensorflow.org/datasets/catalog/emnist#emnistletters}{emnist/letters} &
\citep{Cohen2017EMNISTEM} & 
\href{https://creativecommons.org/licenses/by/4.0/}{Creative Commons Attribution 4.0 License}
\\
\href{https://www.tensorflow.org/datasets/catalog/kmnist}{kmnist} &
\citep{Clanuwat2018DeepLF} &
\href{https://creativecommons.org/licenses/by-sa/4.0/}{Attribution-ShareAlike 4.0 International}
\\
\href{https://www.tensorflow.org/datasets/catalog/mnist}{mnist} &
\citep{LeCun1998GradientbasedLA} &
\href{https://creativecommons.org/licenses/by/4.0/}{Creative Commons Attribution 4.0 License}
\\
\href{https://www.tensorflow.org/datasets/catalog/omniglot}{omniglot} &
\citep{Lake2015HumanlevelCL} &
\href{https://github.com/brendenlake/omniglot/blob/master/LICENSE}{The MIT License}
\\
\href{https://www.tensorflow.org/datasets/catalog/cmaterdb#cmaterdbbangla_default_config}{cmaterdb/bangla} &
\citep{Das2012AGA,Das2012ASF} &
\href{http://www.apache.org/licenses/LICENSE-2.0}{Apache License 2.0}
\\
\href{https://www.tensorflow.org/datasets/catalog/cmaterdb#cmaterdbdevanagari}{cmaterdb/devanagari} &
\citep{Das2012AGA,Das2012ASF} &
\href{http://www.apache.org/licenses/LICENSE-2.0}{Apache License 2.0}
 \\
\href{https://www.tensorflow.org/datasets/catalog/cmaterdb#cmaterdbtelugu}{cmaterdb/telugu} &
\citep{Das2012AGA,Das2012ASF} &
\href{http://www.apache.org/licenses/LICENSE-2.0}{Apache License 2.0}
 \\
 
 \midrule
\multicolumn{3}{c}{\textbf{Visual Domain Decathlon Benchmark}} \\

\href{https://www.tensorflow.org/datasets/catalog/visual_domain_decathlon#visual_domain_decathlonaircraft_default_config}{aircraft} &
\citep{hakanbilensylvestrerebuffitomasjakab2017} & 
\href{https://creativecommons.org/licenses/by/4.0/}{Creative Commons Attribution 4.0 License}
\\
\href{https://www.tensorflow.org/datasets/catalog/visual_domain_decathlon#visual_domain_decathloncifar100}{cifar100} &
\citep{hakanbilensylvestrerebuffitomasjakab2017} & 
\href{https://creativecommons.org/licenses/by/4.0/}{Creative Commons Attribution 4.0 License}
\\
\href{https://www.tensorflow.org/datasets/catalog/visual_domain_decathlon#visual_domain_decathlondaimlerpedcls}{daimlerpedcls} &
\citep{hakanbilensylvestrerebuffitomasjakab2017} & 
\href{https://creativecommons.org/licenses/by/4.0/}{Creative Commons Attribution 4.0 License}
\\
\href{https://www.tensorflow.org/datasets/catalog/visual_domain_decathlon#visual_domain_decathlondtd}{dtd} &
\citep{hakanbilensylvestrerebuffitomasjakab2017} & 
\href{https://creativecommons.org/licenses/by/4.0/}{Creative Commons Attribution 4.0 License}
\\
\href{https://www.tensorflow.org/datasets/catalog/visual_domain_decathlon#visual_domain_decathlongtsrb}{gtsrb} &
\citep{hakanbilensylvestrerebuffitomasjakab2017} & 
\href{https://creativecommons.org/licenses/by/4.0/}{Creative Commons Attribution 4.0 License}
\\
\href{https://www.tensorflow.org/datasets/catalog/visual_domain_decathlon#visual_domain_decathlonimagenet12}{imagenet12} &
\citep{hakanbilensylvestrerebuffitomasjakab2017} & 
\href{https://creativecommons.org/licenses/by/4.0/}{Creative Commons Attribution 4.0 License}
\\
\href{https://www.tensorflow.org/datasets/catalog/visual_domain_decathlon#visual_domain_decathlonomniglot}{omniglot} &
\citep{hakanbilensylvestrerebuffitomasjakab2017} & 
\href{https://creativecommons.org/licenses/by/4.0/}{Creative Commons Attribution 4.0 License}
\\
\href{https://www.tensorflow.org/datasets/catalog/visual_domain_decathlon#visual_domain_decathlonsvhn}{svhn} &
\citep{hakanbilensylvestrerebuffitomasjakab2017} & 
\href{https://creativecommons.org/licenses/by/4.0/}{Creative Commons Attribution 4.0 License}
\\
\href{https://www.tensorflow.org/datasets/catalog/visual_domain_decathlon#visual_domain_decathlonucf101}{ucf101} &
\citep{hakanbilensylvestrerebuffitomasjakab2017} & 
\href{https://creativecommons.org/licenses/by/4.0/}{Creative Commons Attribution 4.0 License}
\\
\href{https://www.tensorflow.org/datasets/catalog/visual_domain_decathlon#visual_domain_decathlonvgg-flowers}{vgg-flowers} &
\citep{hakanbilensylvestrerebuffitomasjakab2017} & 
\href{https://creativecommons.org/licenses/by/4.0/}{Creative Commons Attribution 4.0 License}
\\
    \bottomrule
  \end{tabular}
\end{table}

\clearpage
\section{Experimental details}
This section reports additinoal details 
of the experiments discussed in the paper.

\begin{table}[h]
\caption{
Per task details of the models comparison on the Multitask Character Classification Benchmark reported in Table~\ref{table:chars}.
For each task is reported the mean and standard deviation of the test and validation top 1 accuracy computed over the experiment repetitions.
}
\label{table:chars-extra}
\centering
\small
\setlength\tabcolsep{3pt}
\hspace*{-1.8pt}
\begin{tabular}{lcccccccc}
\toprule
 
Model
& 
\rotatebox{90}{\href{https://www.tensorflow.org/datasets/catalog/emnist\#emnistdigits}{emnist/}} 
\rotatebox{90}{\href{https://www.tensorflow.org/datasets/catalog/emnist\#emnistdigits}{digits}} 
&
\rotatebox{90}{\href{https://www.tensorflow.org/datasets/catalog/emnist\#emnistletters}{emnist/}}
\rotatebox{90}{\href{https://www.tensorflow.org/datasets/catalog/emnist\#emnistletters}{letters}}
&
\rotatebox{90}{\href{https://www.tensorflow.org/datasets/catalog/kmnist}{kmnist}}
&
\rotatebox{90}{\href{https://www.tensorflow.org/datasets/catalog/mnist}{mnist}}
&
\rotatebox{90}{\href{https://www.tensorflow.org/datasets/catalog/omniglot}{omniglot}}
&
\rotatebox{90}{\href{https://www.tensorflow.org/datasets/catalog/cmaterdb\#cmaterdbbangla_default_config}{cmaterdb/}}
\rotatebox{90}{\href{https://www.tensorflow.org/datasets/catalog/cmaterdb\#cmaterdbbangla_default_config}{bangla}}
&
\rotatebox{90}{\href{https://www.tensorflow.org/datasets/catalog/cmaterdb\#cmaterdbdevanagari}{cmaterdb/}}
\rotatebox{90}{\href{https://www.tensorflow.org/datasets/catalog/cmaterdb\#cmaterdbdevanagari}{devanagari}}
&
\rotatebox{90}{\href{https://www.tensorflow.org/datasets/catalog/cmaterdb\#cmaterdbtelugu}{cmaterdb/}}
\rotatebox{90}{\href{https://www.tensorflow.org/datasets/catalog/cmaterdb\#cmaterdbtelugu}{telugu}}

\\

\midrule
 & \multicolumn{8}{c}{\textbf{Test accuracy}}
\\

\href{https://youtu.be/LIvOCmF1aRk}{Multi-head}
&  37.1\plmi2.1 & 12.1\plmi2.2 &  20.7\plmi1.0 &  38.0\plmi1.4 &  0.3\plmi0.3 &  18.8\plmi1.5 &  18.6\plmi3.3 &  16.5\plmi3.0 
\\

\href{https://youtu.be/_fikzbxS_ZY}{Full fine-tuning}
&  98.2\plmi0.1 &  87.4\plmi0.1 &  91.5\plmi0.3 &  98.2\plmi0.1 &  49.1\plmi1.0 &  82.4\plmi0.8 &  63.4\plmi2.1 &  81.3\plmi1.1
\\
\href{https://youtu.be/R3ETGxo9CWE}{R.adapter dim=512}
&  98.2\plmi0.1 &  86.5\plmi0.3 &  89.5\plmi0.1 &  98.2\plmi0.1 &  47.9\plmi0.9 &  85.2\plmi0.7 &  67.8\plmi1.3 &  82.6\plmi0.7
\\

\href{https://youtu.be/CQdeP1mpr-8}{muNet w/o a.tuning}
&  98.1\plmi0.1 &  87.9\plmi0.3 &  91.8\plmi0.3 &  98.1\plmi0.2 &  66.6\plmi1.8 &  90.7\plmi0.6 &  83.2\plmi1.6 &  92.1\plmi1.0
\\

\href{https://youtu.be/Ld9gfmJT6Ig}{muNet scale=0.3}
&  95.8\plmi1.5 &  80.9\plmi4.8 &  80.2\plmi7.6 &  95.5\plmi1.9 &  57.0\plmi5.1 &  85.0\plmi3.2 &  77.8\plmi5.7 &  84.0\plmi4.8 
\\
\href{https://youtu.be/-xOl3lJV4fw}{muNet}
&  98.4\plmi0.4 &  89.3\plmi1.5 &  92.7\plmi1.8 &  98.6\plmi0.4 &  78.8\plmi5.7 &  93.9\plmi1.2 &  85.8\plmi2.0 &  93.9\plmi1.8
\\

\midrule
& \multicolumn{8}{c}{\textbf{Validation accuracy}}
\\

\href{https://youtu.be/LIvOCmF1aRk}{Multi-head}
&  37.3\plmi1.7 &  16.9\plmi0.6 &  26.7\plmi1.1 &  36.8\plmi1.2 &  1.0\plmi0.2 &  19.4\plmi1.4 &  17.3\plmi1.6 &  18.1\plmi1.0
\\

\href{https://youtu.be/_fikzbxS_ZY}{Full fine-tuning}
 &  98.2\plmi0.1 &  88.3\plmi0.2 &  96.9\plmi0.2 &  98.2\plmi0.1 &  48.2\plmi0.8 &  83.0\plmi1.1 &  62.9\plmi1.2 &  80.6\plmi0.6
\\
\href{https://youtu.be/R3ETGxo9CWE}{R.adapter dim=512}
&  98.3\plmi0.0 &  87.8\plmi0.1 &  96.1\plmi0.1 &  98.2\plmi0.1 &  47.3\plmi0.7 &  89.0\plmi0.2 &  73.1\plmi0.8 &  82.6\plmi1.2
\\

\href{https://youtu.be/CQdeP1mpr-8}{muNet w/o a.tuning}
&  98.1\plmi0.1 &  89.4\plmi0.3 &  96.8\plmi0.2 &  97.9\plmi0.2 &  63.2\plmi1.8 &  94.3\plmi1.2 &  86.2\plmi1.6 &  93.9\plmi2.2
\\

\href{https://youtu.be/Ld9gfmJT6Ig}{muNet scale=0.3}
&  96.0\plmi1.3 &  83.4\plmi4.7 &  90.6\plmi3.8 &  95.4\plmi2.1 &  56.6\plmi5.4 &  86.9\plmi3.7 &  81.4\plmi7.5 &  86.4\plmi6.1
\\
\href{https://youtu.be/-xOl3lJV4fw}{muNet}
&  98.4\plmi0.3 &  90.8\plmi1.7 &  97.4\plmi0.7 &  98.4\plmi0.4 &  76.7\plmi5.7 &  96.5\plmi1.2 &  92.6\plmi2.2 &  95.8\plmi2.4
\\

\bottomrule
\end{tabular}
\end{table}


\begin{table}[h]
  \caption{
  Per task details of the models comparison on the Visual Domain Decathlon Benchmark reported in Table~\ref{table:deca}.
For each task is reported the mean and standard deviation of the test and validation top 1 accuracy computed over the experiment repetitions.
}
  \label{table:deca-extra}
\centering
\small
\setlength\tabcolsep{1.5pt}
\begin{tabular}{lcccccccccc}
    \toprule
Model
& 
\rotatebox{90}{\href{https://www.tensorflow.org/datasets/catalog/visual_domain_decathlon\#visual_domain_decathlonimagenet12}{imagenet12}}
&
\rotatebox{90}{\href{https://www.tensorflow.org/datasets/catalog/visual_domain_decathlon\#visual_domain_decathlonsvhn}{svhn}}
&
\rotatebox{90}{\href{https://www.tensorflow.org/datasets/catalog/visual_domain_decathlon\#visual_domain_decathloncifar100}{cifar100}}
&
\rotatebox{90}{\href{https://www.tensorflow.org/datasets/catalog/visual_domain_decathlon\#visual_domain_decathlongtsrb}{gtsrb}}
&
\rotatebox{90}{\href{https://www.tensorflow.org/datasets/catalog/visual_domain_decathlon\#visual_domain_decathlondaimlerpedcls}{daimlerpedcls}}
&
\rotatebox{90}{\href{https://www.tensorflow.org/datasets/catalog/visual_domain_decathlon\#visual_domain_decathlonomniglot}{omniglot}}
&
\rotatebox{90}{\href{https://www.tensorflow.org/datasets/catalog/visual_domain_decathlon\#visual_domain_decathlonucf101}{ucf101}}
&
\rotatebox{90}{\href{https://www.tensorflow.org/datasets/catalog/visual_domain_decathlon\#visual_domain_decathlonaircraft_default_config}{aircraft}}
&
\rotatebox{90}{\href{https://www.tensorflow.org/datasets/catalog/visual_domain_decathlon\#visual_domain_decathlondtd}{dtd}}
&
\rotatebox{90}{\href{https://www.tensorflow.org/datasets/catalog/visual_domain_decathlon\#visual_domain_decathlonvgg-flowers}{vgg-flowers}}

\\
\midrule
& \multicolumn{10}{c}{\textbf{Test accuracy}}
\\
\href{https://youtu.be/7UgPZYgh53U}{Multi-head}
&  34.6\plmi0.1 &  40.5\plmi0.2 &  55.9\plmi0.1 &  73.3\plmi0.2 &  91.5\plmi0.1 &  22.1\plmi0.3 &  39.6\plmi0.7 &  17.3\plmi0.5 &  48.4\plmi0.6 &  74.1\plmi0.7 
\\

\href{https://youtu.be/vNo-j150nA0}{Unfreeze above 1st}
&  73.8\plmi0.2 &  93.8\plmi0.2 &  88.7\plmi0.2 &  99.9\plmi0.0 &  99.8\plmi0.1 &  81.6\plmi0.4 &  78.6\plmi1.1 &  41.0\plmi0.4 &  62.8\plmi0.2 &  93.9\plmi0.3
\\

\href{https://youtu.be/BK7AW95ii4s}{Full fine-tuning}
&  73.2\plmi0.1 &  93.5\plmi0.1 &  87.5\plmi0.4 &  99.9\plmi0.1 &  99.8\plmi0.1 &  81.5\plmi0.1 &  78.6\plmi0.1 &  41.2\plmi1.0 &  60.2\plmi0.5 &  92.8\plmi0.6
\\

\href{https://youtu.be/oWiniz6F2Lw}{R.adapters dim=512}
&  69.5\plmi0.1 &  93.7\plmi0.1 &  88.7\plmi0.2 &  99.9\plmi0.0 &  99.9\plmi0.0 &  81.2\plmi0.2 &  76.0\plmi0.2 &  40.8\plmi0.2 &  61.4\plmi1.1 &  94.4\plmi0.2
 \\

\href{https://youtu.be/P0SBFOuyj0s}{muNet w/o a.tuning}
&  71.6\plmi0.4 &  93.8\plmi0.2 &  89.7\plmi0.2 &  99.9\plmi0.1 &  99.9\plmi0.0 &  81.9\plmi0.6 &  80.3\plmi1.9 &  43.2\plmi0.4 &  66.1\plmi0.6 &  96.3\plmi0.4
\\

\href{https://youtu.be/THyc5lUC_-w}{muNet scale=0.3}
&  68.1\plmi0.7 &  92.3\plmi1.4 &  88.4\plmi1.2 &  98.2\plmi0.8 &  99.7\plmi0.2 &  76.9\plmi1.6 &  72.7\plmi2.1 &  40.2\plmi2.9 &  64.7\plmi0.6 &  94.6\plmi1.5
\\

\href{https://youtu.be/2scExBaHweY}{muNet}
&  74.3\plmi2.3 &  94.6\plmi0.4 &  90.2\plmi0.9 &  99.9\plmi0.0 &  99.9\plmi0.1 &  84.0\plmi1.7 &  79.7\plmi1.9 &  47.2\plmi0.8 &  65.9\plmi1.1 &  96.3\plmi0.8
\\

\midrule
& \multicolumn{10}{c}{\textbf{Validation accuracy}}
\\
\href{https://youtu.be/7UgPZYgh53U}{Multi-head}
&  36.6\plmi0.1 &  41.6\plmi0.3 &  56.7\plmi0.2 &  73.9\plmi0.1 &  90.5\plmi0.1 &  22.6\plmi0.2 &  40.2\plmi0.2 &  18.3\plmi0.1 &  48.1\plmi0.1 &  77.3\plmi0.4
\\

\href{https://youtu.be/vNo-j150nA0}{Unfreeze above 1st}
&  74.6\plmi0.3 &  94.9\plmi0.0 &  89.5\plmi0.2 &  100\plmi0.0 &  100\plmi0.0 &  82.6\plmi0.1 &  78.6\plmi0.3 &  39.7\plmi0.2 &  62.0\plmi0.3 &  93.8\plmi0.2
\\

\href{https://youtu.be/BK7AW95ii4s}{Full fine-tuning}
&  74.0\plmi0.2 &  94.2\plmi0.0 &  88.4\plmi0.1 &  100\plmi0.0 &  99.9\plmi0.0 &  82.6\plmi0.1 &  77.2\plmi0.4 &  40.7\plmi0.3 &  60.1\plmi0.4 &  93.2\plmi0.1
\\

\href{https://youtu.be/oWiniz6F2Lw}{R.adapters dim=512}
&  70.2\plmi0.2 &  94.3\plmi0.1 &  90.4\plmi0.0 &  100\plmi0.0 &  100\plmi0.0 &  82.3\plmi0.1 &  75.2\plmi0.1 &  39.7\plmi0.1 &  62.9\plmi0.2 &  94.1\plmi0.0 
\\

\href{https://youtu.be/P0SBFOuyj0s}{muNet w/o a.tuning}
&  72.9\plmi0.5 &  94.5\plmi0.2 &  90.7\plmi0.2 &  100\plmi0.0 &  100\plmi0.0 &  83.5\plmi0.1 &  79.0\plmi0.4 &  43.0\plmi1.0 &  65.8\plmi0.3 &  97.0\plmi0.2
\\

\href{https://youtu.be/THyc5lUC_-w}{muNet scale=0.3}
&  69.8\plmi0.8 &  92.8\plmi1.1 &  89.4\plmi0.9 &  98.0\plmi0.8 &  99.8\plmi0.1 &  77.4\plmi1.8 &  71.9\plmi1.4 &  39.3\plmi3.1 &  63.7\plmi1.0 &  95.2\plmi0.6
\\

\href{https://youtu.be/2scExBaHweY}{muNet}
&  75.8\plmi2.2 &  95.6\plmi0.4 &  91.4\plmi0.7 &  100\plmi0.0 &  100\plmi0.0 &  84.9\plmi1.8 &  78.9\plmi1.4 &  47.0\plmi1.4 &  65.6\plmi0.4 &  96.7\plmi0.4
\\
    \bottomrule
  \end{tabular}
\end{table}

\begin{figure}[h]
\centering
\includegraphics[width=.8\linewidth]{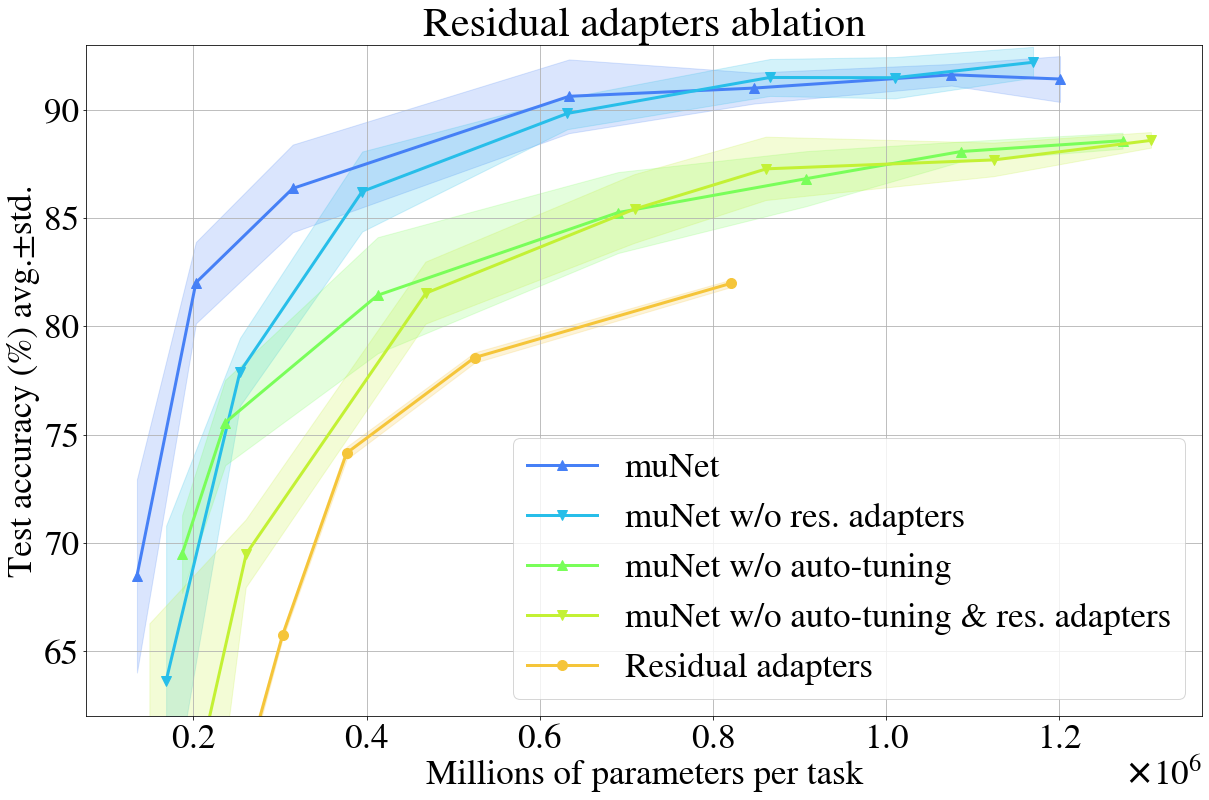}
\caption{
Residual adapters ablation experiments on the  Multitask Character Classification Benchmark.
The muNet experiments are repeated without the layer insertion action that is needed to insert residual adapters in the generated architectures.
The residual adapter ablation is repeated also for the "muNet without auto-tuning" configuration.
All other aspects of the experiment are unchanged, such as: scale factors considered: \{0.02, 0.3, 0.7, 0.9, 0.95, 0.98, 1\}, 5 repetitions for each scale factor, and equivalent train budget for each experiment.
Curves plot top 1 test accuracy averaged across the 5 repetitions, and the shaded area represents standard deviation.
Notice that for scale factors above 0.9 the corresponding curves with/without residual adapters are equivalent within noise, while for lower scale factors the absence of residual adapters leads to a significant worsening of the achieved quality/size trade-offs.
For reference, the performance achieved by the residual adapters baseline is also plotted.
Notice that muNet with neither residual adapters nor auto-tuning still outperforms the residual adapters baseline across the whole range.
}
\label{fig:ablate-ra}
\end{figure}


\begin{figure}[h]
\centering
\includegraphics[width=1.\linewidth]{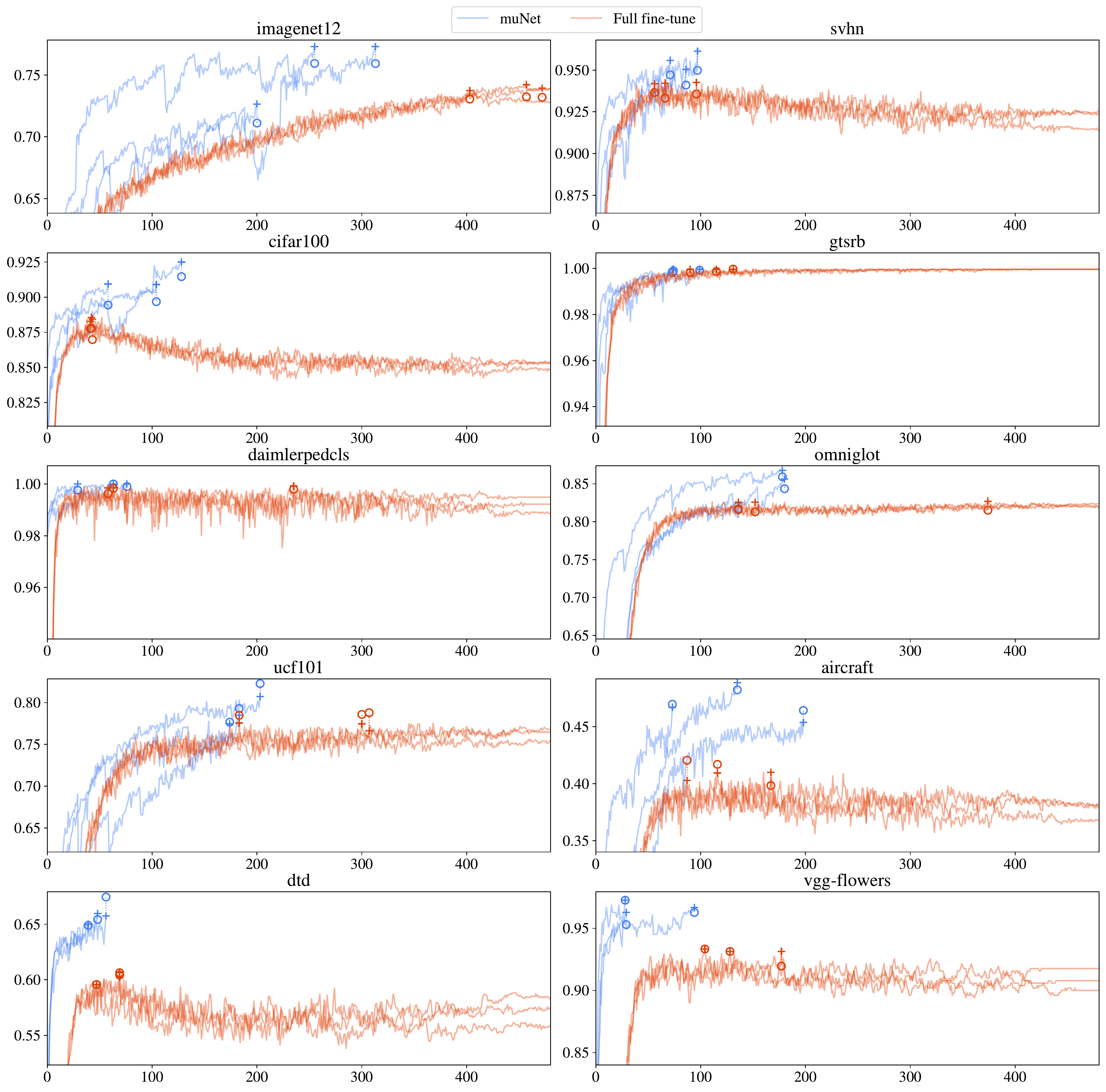}
\caption{
Validation quality measured during training of the best model achieved by each of the 3 repetitions of muNet and full fine-tune baseline models on each of the Visual Domain Decathlon Benchmark tasks.
The vertical axis measures the top 1 accuracy over the validation set.
The horizontal axis measures the number of training-validation cycles.
The cross markers, {\tiny$+$}, highlight the validation accuracy achieved by the best models.
The circle markers, $\circ$, display the test accuracy achieved by the best models.
Curves for the muNet repetitions are shorter since they include only the training-validation cycles performed by the model (and its ancestors) that results being the best at the end of the final active task cycle.
We observe that muNet models archive significant gains on the smaller tasks, this can be expected as the smaller tasks have fewer data to train on and can benefit the most from transfer of knowledge.
But also we observe significant gains on larger tasks like imagenet12 or cifar100.
Baseline models display lower variance.
}
\label{fig:crvs}
\end{figure}

\begin{figure}[h]
\centering
\includegraphics[width=1.\linewidth]{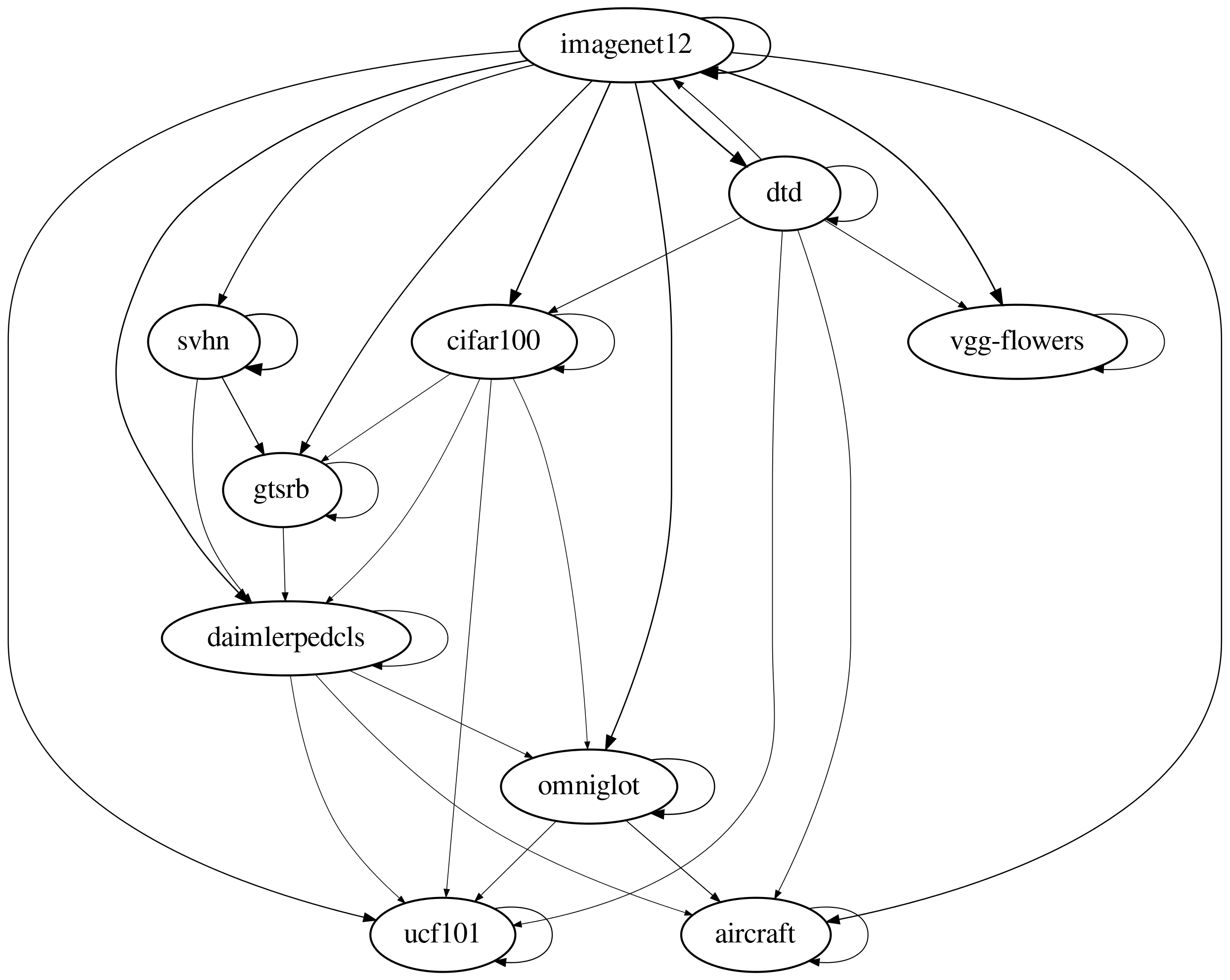}
\caption{
Graph representing the knowledge flow for the best models generated by muNet for the Visual Domain Decathlon Benchmark tasks.
Each node represents a task.
Each directional edge from a source task to a target task represents the usage by the target task best model of knowledge/parameters introduced by the source task.
Knowledge transfer is quantified as the fraction of training cycles performed on the source tasks by the parameters included the best model for the target task.
The thicker the edge and arrow, the more training has been performed on the source task.
Edges starting and ending on the same node represent the fraction of the total training that has been performed on the target task itself.
Specialized tasks with small dataset, such as vgg-flowers, aircraft and ucf101, reuse knowledge from multiple tasks, but no other task reuses parameters fine-tuned by them.
And vice versa, generic tasks with large datasets, such as imagenet12, contribute to all other tasks, but their best model is mostly trained on their own dataset.
}
\label{fig:flow-deca}
\end{figure}

\begin{figure}[h]
\centering
\includegraphics[width=1.\linewidth]{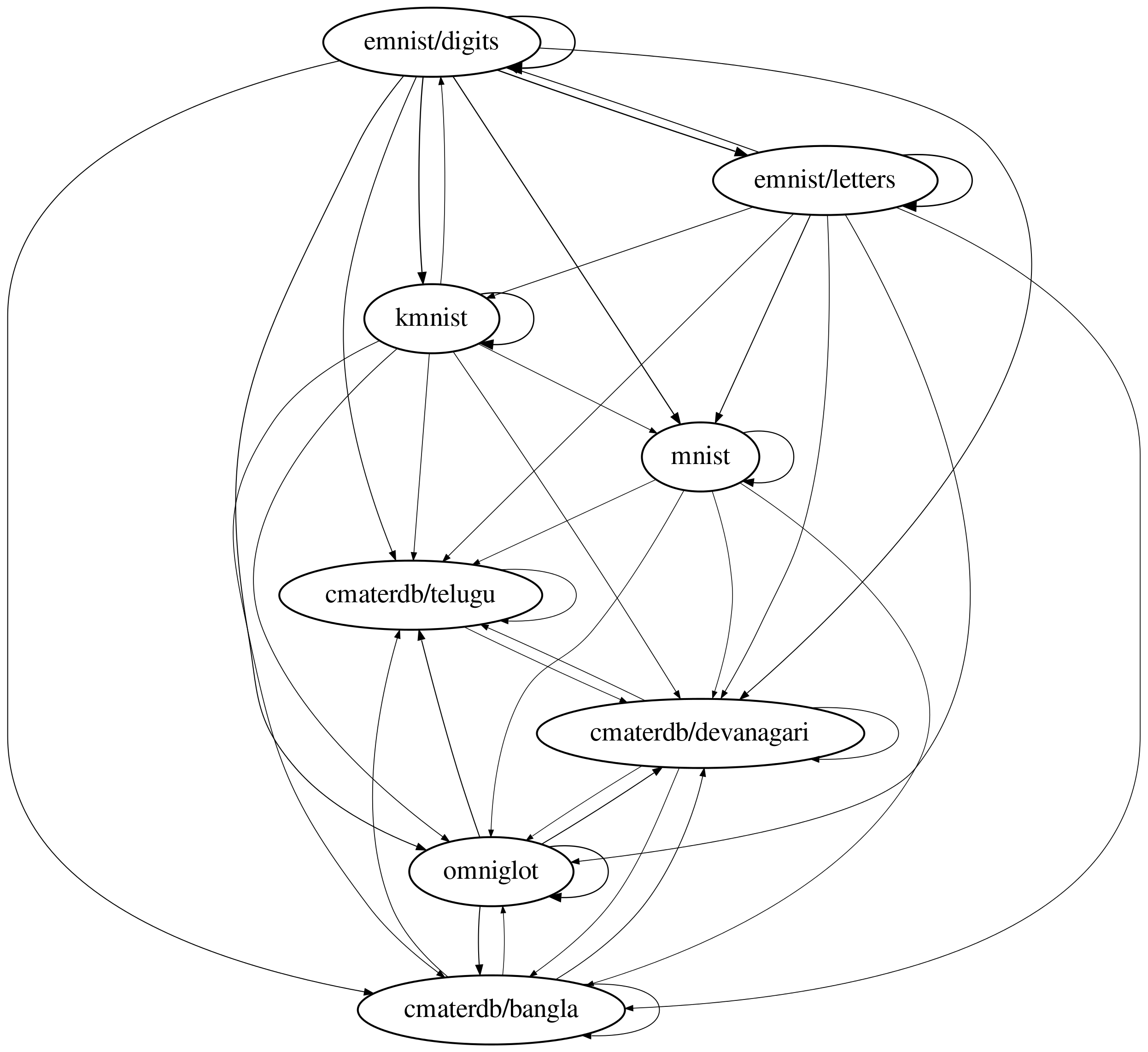}
\caption{
Graph representing the knowledge flow for the best models generated by muNet for the Multitask Character Classification tasks.
Each node represents a task.
Each directional edge from a source task to a target task represents the usage by the target task best model of knowledge/parameters introduced by the source task.
Knowledge transfer is quantified as the fraction of training cycles performed on the source tasks by the parameters included the best model for the target task.
The thicker the edge and arrow, the more training has been performed on the source task.
Edges starting and ending on the same node represent the fraction of the total training that has been performed on the target task itself.
Compared to the results for Visual Domain Decathlon (see Figure~\ref{fig:flow-deca}),
we notice a more distributed knowledge transfer, that can be expected within a set of tasks of similar domain.
Also we notice the same pattern that tasks with more training data mostly rely on the knowledge learned from their own datasets.
In this case the larger tasks are emnist/digits and emnist/letter, that train mostly on their own data with also a significant exchange on knowledge between the two tasks since they are very related and have images in similar format.
Also interesting to notice that the 4 smaller tasks (omniglot and the 3 cmarterdb tasks) transfer knowledge from at least 7 of the 8 tasks.
}
\label{fig:flow-char}
\end{figure}

\clearpage

\begin{figure}[t]
\centering
\begin{minipage}{1.0\textwidth}
  \centering
  \includegraphics[width=1.0\linewidth]{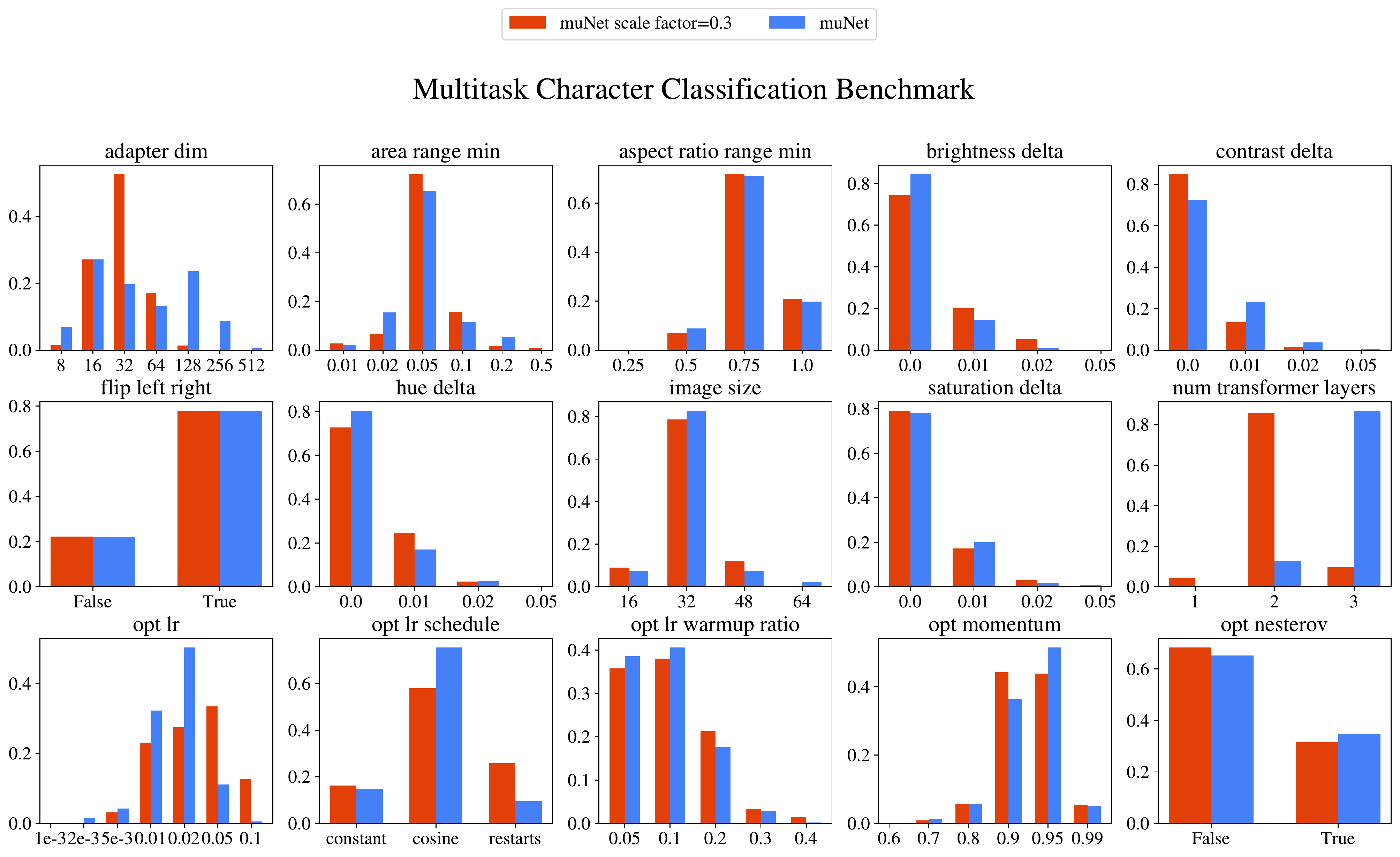}
\end{minipage}
\vspace{0.4cm}

\begin{minipage}{1.0\textwidth}
  \centering
  \includegraphics[width=1.0\linewidth]{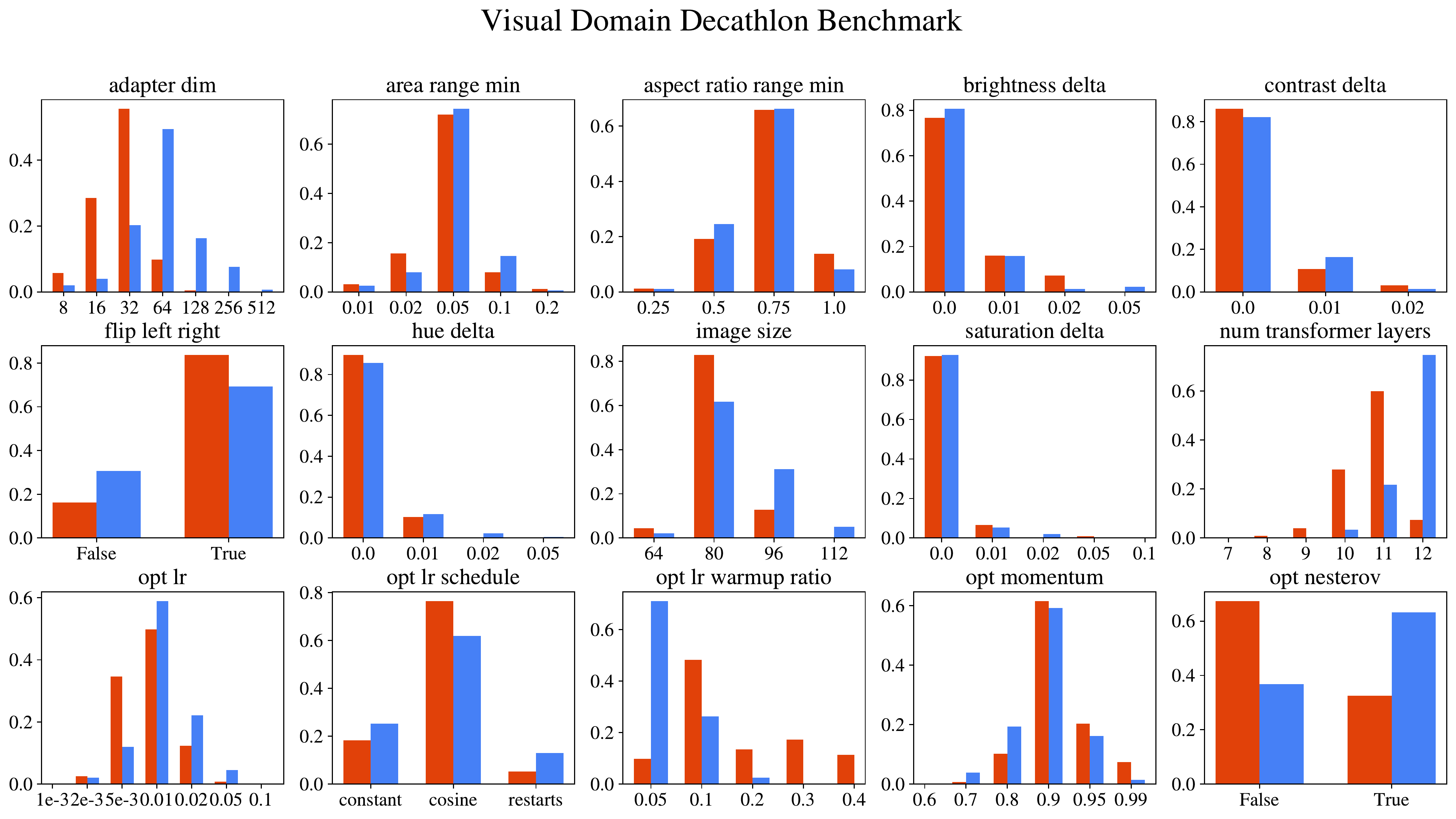}
\end{minipage}
\caption{Distributions of the hyperparameters sampled for the models generated with the "muNet scale factor=0.3" and muNet (scale factor=1) configurations for the Multitask Character Classification Benchmark (top) and Visual Domain Decathlon Benchmark (bottom), aggregated across experiment repetitions.
Notice that the configuration with scale factor 0.3 converges to distributions that allow to achieve a different quality/size trade-off in accordance with the size penalty integrated in the scoring function, such as: smaller adapters dimension and less transformer layers.
}
\label{fig:hps}
\end{figure}

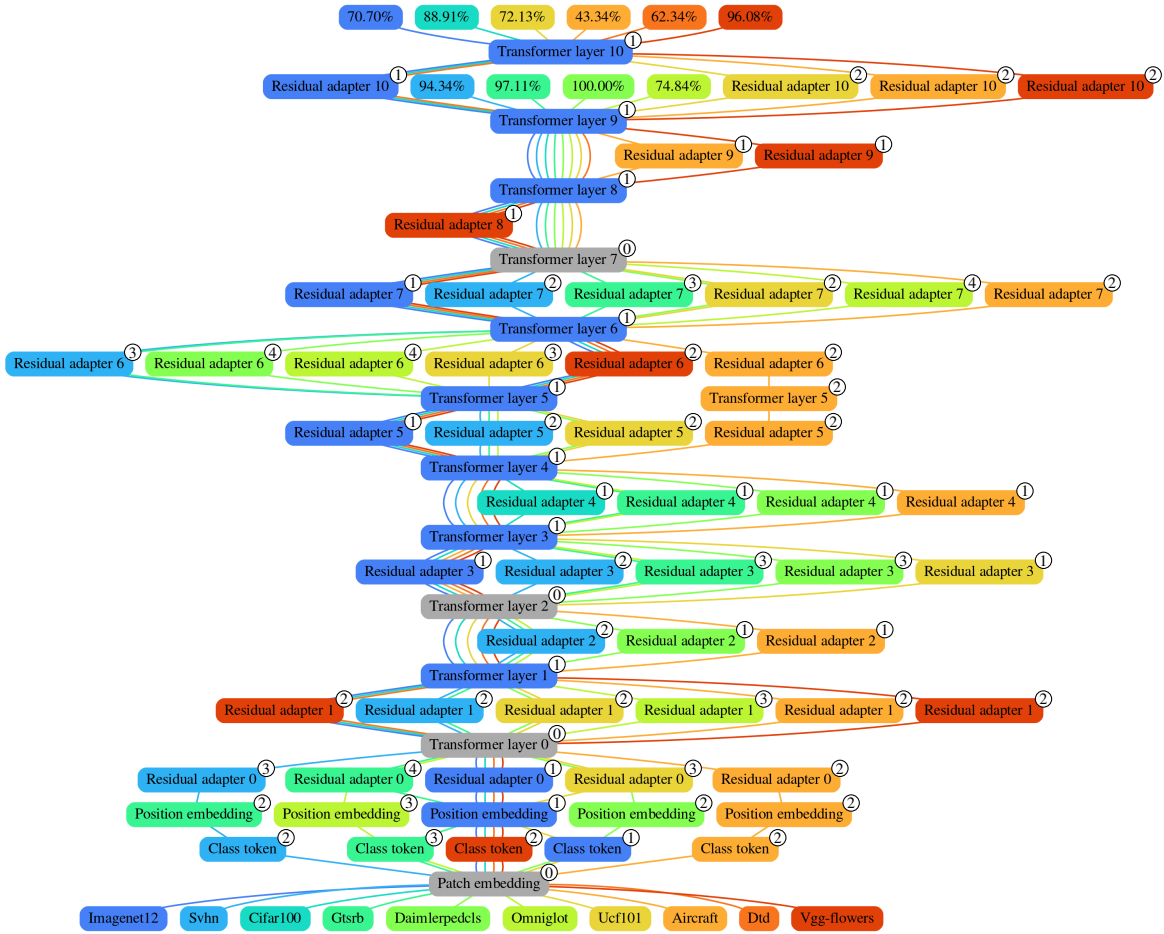
\begin{figure}[t]
\hspace*{-24.155pt}
\centering
\begin{tikzpicture}
\node (image) at (0,0) {\includegraphics[width=1.10\linewidth]{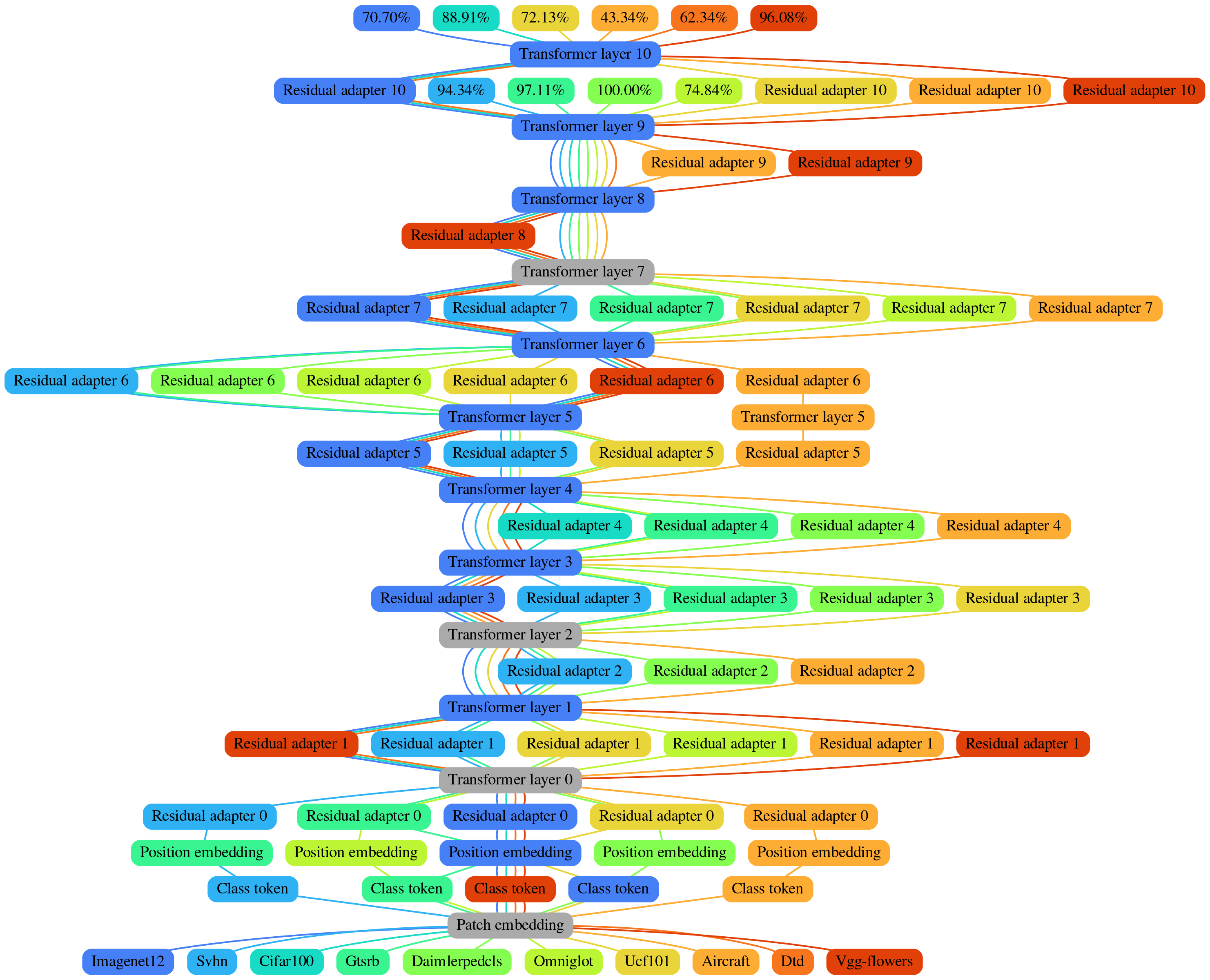}};
\node[stuff_fill] at (20.25pt, 162pt) {\tiny 1};
\node[stuff_fill] at (-68.5pt, 148.875pt) {\tiny 1};
\node[stuff_fill] at (105.5pt, 148.875pt) {\tiny 2};
\node[stuff_fill] at (161.25pt, 148.875pt) {\tiny 2};
\node[stuff_fill] at (217pt, 148.875pt) {\tiny 2};
\node[stuff_fill] at (18pt, 135.75pt) {\tiny 1};
\node[stuff_fill] at (62pt, 122.625pt) {\tiny 1};
\node[stuff_fill] at (115pt, 122.625pt) {\tiny 1};
\node[stuff_fill] at (18pt, 109.5pt) {\tiny 1};
\node[stuff_fill] at (-25pt, 96.375pt) {\tiny 1};
\node[stuff_fill] at (18pt, 83.25pt) {\tiny 0};
\node[stuff_fill] at (-63pt, 70.125pt) {\tiny 1};
\node[stuff_fill] at (-10.1pt, 70.125pt) {\tiny 2};
\node[stuff_fill] at (42.8pt, 70.125pt) {\tiny 3};
\node[stuff_fill] at (95.7pt, 70.125pt) {\tiny 2};
\node[stuff_fill] at (148.6pt, 70.125pt) {\tiny 4};
\node[stuff_fill] at (201.5pt, 70.125pt) {\tiny 2};
\node[stuff_fill] at (18pt, 57pt) {\tiny 1};
\node[stuff_fill] at (-169pt, 43.875pt) {\tiny 3};
\node[stuff_fill] at (-116pt, 43.875pt) {\tiny 4};
\node[stuff_fill] at (-63pt, 43.875pt) {\tiny 4};
\node[stuff_fill] at (-10pt, 43.875pt) {\tiny 3};
\node[stuff_fill] at (43pt, 43.875pt) {\tiny 2};
\node[stuff_fill] at (96pt, 43.875pt) {\tiny 2};
\node[stuff_fill] at (-8.5pt, 30.75pt) {\tiny 1};
\node[stuff_fill] at (97.5pt, 30.75pt) {\tiny 2};
\node[stuff_fill] at (-63pt, 17.625pt) {\tiny 1};
\node[stuff_fill] at (-10pt, 17.625pt) {\tiny 2};
\node[stuff_fill] at (43pt, 17.625pt) {\tiny 2};
\node[stuff_fill] at (96pt, 17.625pt) {\tiny 2};
\node[stuff_fill] at (-8.5pt, 4.5pt) {\tiny 1};
\node[stuff_fill] at (9.5pt, -8.625pt) {\tiny 1};
\node[stuff_fill] at (62.5pt, -8.625pt) {\tiny 1};
\node[stuff_fill] at (115.5pt, -8.625pt) {\tiny 1};
\node[stuff_fill] at (168.5pt, -8.625pt) {\tiny 1};
\node[stuff_fill] at (-8.5pt, -21.75pt) {\tiny 1};
\node[stuff_fill] at (-36.5pt, -34.875pt) {\tiny 1};
\node[stuff_fill] at (16.5pt, -34.875pt) {\tiny 2};
\node[stuff_fill] at (69.5pt, -34.875pt) {\tiny 3};
\node[stuff_fill] at (122.5pt, -34.875pt) {\tiny 3};
\node[stuff_fill] at (175.5pt, -34.875pt) {\tiny 1};
\node[stuff_fill] at (-8.5pt, -48pt) {\tiny 0};
\node[stuff_fill] at (9.5pt, -61.125pt) {\tiny 2};
\node[stuff_fill] at (62.5pt, -61.125pt) {\tiny 1};
\node[stuff_fill] at (115.5pt, -61.125pt) {\tiny 1};
\node[stuff_fill] at (-8.5pt, -74.25pt) {\tiny 1};
\node[stuff_fill] at (-89pt, -87.375pt) {\tiny 2};
\node[stuff_fill] at (-36.1pt, -87.375pt) {\tiny 2};
\node[stuff_fill] at (16.8pt, -87.375pt) {\tiny 2};
\node[stuff_fill] at (69.7pt, -87.375pt) {\tiny 3};
\node[stuff_fill] at (122.6pt, -87.375pt) {\tiny 2};
\node[stuff_fill] at (175.5pt, -87.375pt) {\tiny 2};
\node[stuff_fill] at (-8.5pt, -100.5pt) {\tiny 0};
\node[stuff_fill] at (-118.5pt, -113.625pt) {\tiny 3};
\node[stuff_fill] at (-63pt, -113.625pt) {\tiny 4};
\node[stuff_fill] at (-10pt, -113.625pt) {\tiny 1};
\node[stuff_fill] at (43pt, -113.625pt) {\tiny 3};
\node[stuff_fill] at (98.5pt, -113.625pt) {\tiny 2};
\node[stuff_fill] at (-120pt, -126.75pt) {\tiny 2};
\node[stuff_fill] at (-64.5pt, -126.75pt) {\tiny 3};
\node[stuff_fill] at (-8.5pt, -126.75pt) {\tiny 1};
\node[stuff_fill] at (47pt, -126.75pt) {\tiny 2};
\node[stuff_fill] at (102.75pt, -126.75pt) {\tiny 2};
\node[stuff_fill] at (-111pt, -139.875pt) {\tiny 2};
\node[stuff_fill] at (-55pt, -139.875pt) {\tiny 3};
\node[stuff_fill] at (-17.25pt, -139.875pt) {\tiny 2};
\node[stuff_fill] at (19.5pt, -139.875pt) {\tiny 1};
\node[stuff_fill] at (75pt, -139.875pt) {\tiny 2};
\node[stuff_fill] at (-11.5pt, -153pt) {\tiny 0};
\end{tikzpicture}
\caption{Model graph representing the multitask network for the Visual Domain Decathlon Benchmark tasks displayed on the bottom nodes,
generated by \href{https://youtu.be/THyc5lUC_-w}{muNet with scale factor=0.3} (see Section~\ref{subsection:evo}).
This model was generated by the experiment repetition achieving the max average test accuracy across the tasks (see Section~\ref{subsection:deca}).
Each task is identified with a unique color.
Top nodes represent the head layer of each task, and display the validation accuracy for that task.
Each sequence of edges of the same color connecting a task input to its head, defines the layers sequence composing the model for each task.
Internal nodes are represented with the color of the task on which the parameters of the corresponding layer were trained last.
Except for the gray nodes that have not received gradients from any of these 10 tasks and still carry the parameters of the root model that were loaded from a checkpoint of a ViT B/16 pretrained on the imagenet-21k dataset (see Section~\ref{subsection:deca}).
The number of unique tasks each layer have been trained on through the sequence of its ancestors is displayed in the top right corner label, {\small \textcircled{n}}.
}
\label{fig:deca-30}
\end{figure}

\clearpage

\begin{algorithm}
\caption{Pseudocode for the generation of a multitask system for a given list of tasks}
\label{algo}
\begin{algorithmic}[1]
\State List of tasks: $T \gets [t_1, t_2, ..., t_{|T|}]$
\State Set of all the models in the multitask system: $\mathcal{M} \gets \{root\mbox{-}model\}$
\For{$\#task\mbox{-}iterations$}
\For{Active task: $t\in T$}
    \State Seed parent models: $\mathcal{S} \gets \mathcal{M}$
    \State Active population: $\mathcal{A} \gets \{m\ |\  m \in \mathcal{M} \land m$\ trained on $t \}$
    \For{$\#generations$}
        \For{$\#child\mbox{-}models$}

            \State \Comment{Sample parent model}
            \State Parent model: $p \gets $ \textbf{none}
            \If{$\mathcal{S} \not= \emptyset$}
                \State $p \sim Uniform(\mathcal{S})$
                \State $\mathcal{S} \gets \{m\ |\ m \in \mathcal{S} \land m \not= p \}  $
            \Else
                \For{Candidate parent model: $\hat{p} \in [sorted_{score}(\mathcal{A})]$}  
                    \If{$\nicefrac{1}{2}^{\#offsprings(\hat{p})} > x \sim Uniform([0, 1])$}
                        \State $p \gets \hat{p}$
                        \State \textbf{break}
                    \EndIf
                \EndFor
                \If{$p=$ \textbf{none}}
                    \State $p \sim Uniform(\mathcal{A} \cup \mathcal{M})$
                \EndIf
            \EndIf

            \State \Comment{Sample child model}
            \State Set of mutations: $\Delta \gets \{make\mbox{-}trainable\mbox{-}head\}$
            \If{$p$ is not seed parent model}
                \For{Candidate mutation: $\hat{\delta} \in possible\mbox{-}mutations(p)$}
                    \If{$\mu \geq x \sim Uniform([0, 1]) $}
                        \State $\Delta \gets \Delta \cup \{\hat{\delta}\}$ 
                    \EndIf
                \EndFor
            \EndIf
            \State Untrained child model: $c_0 \gets apply\mbox{-}mutations(p, \Delta)$

            \State \Comment{Train child model}
            \State Retained child model: $c \gets$ \textbf{none}
            \State Best score: $score^* \gets \mbox{-}\infty$
            \If{$p $ trained on $ t$}
                \State $score^* \gets score(p)$
            \EndIf
            \For{$i \in [1, ...\ , \#train\mbox{-}cycles]$}
                \State $c_i \gets train(c_{i-1}, min(1\ epoch,\ \#samples\mbox{-}cap))$
                \If{$score(c_i) > score^*$}
                    \State $c \gets c^i$
                    \State $score^* \gets score(c^i)$
                \EndIf
            \EndFor
            \If{$c \not= $ \textbf{none}}
                \State $\mathcal{A} \gets \mathcal{A} \cup \{c\}$
            \EndIf
        \EndFor
    \EndFor
    \State \Comment{Keep only the best model for $t$}
    \State $\mathcal{M} \gets \{ \argmax_{m \in \mathcal{A}} score(m)\} \cup \{m\ |\ m \in \mathcal{M} \land m $ not trained on $ t\}$
\EndFor
\EndFor
\end{algorithmic}
\end{algorithm}
\end{document}